\documentclass[11pt,3p,review,authoryear]{elsarticle}
\usepackage{amssymb}
\usepackage{amsmath}
\usepackage{pifont}
\usepackage{mathtools}
\usepackage{todonotes}
\usepackage{booktabs}
\usepackage{subcaption}
\usepackage{caption}
\usepackage{multirow}

\journal{International Journal of Forecasting}
\biboptions{longnamesfirst}
\newcommand{\plusmark}{\ding{53}}%
\newcommand{\smark}{$\dagger$}%

\makeatletter
\def\ps@pprintTitle{%
 \let\@oddhead\@empty
 \let\@evenhead\@empty
 \def\@oddfoot{}%
 \let\@evenfoot\@oddfoot}
\makeatother

\begin{document}

\begin{frontmatter}

\title{NeuralProphet: Explainable Forecasting at Scale}

\author[su]{Oskar Triebe\corref{cor} \fnref{xnote}
}\fntext[xnote]{Maintainer of Github repository.\\ }
\author[mu]{Hansika Hewamalage}
\author[sk]{Polina Pilyugina}
\author[fb]{\\ Nikolay Laptev}
\author[mu]{Christoph Bergmeir}
\author[su]{Ram Rajagopal}

\address[su]{Stanford University}
\address[fb]{Facebook, Inc.}
\address[mu]{Monash University}
\address[sk]{Skoltech University}

\cortext[cor]{Corresponding author. \textit{Email address:} \texttt{triebe@stanford.edu}}

\begin{abstract}
 
We introduce NeuralProphet, a successor to Facebook Prophet, which set an industry standard for explainable, scalable, and user-friendly forecasting frameworks.
With the proliferation of time series data, explainable forecasting remains a challenging task for business and operational decision making.
Hybrid solutions are needed to bridge the gap between interpretable classical methods and scalable deep learning models. We view Prophet as a precursor to such a solution. However, Prophet lacks local context, which is essential for forecasting the near-term future and is challenging to extend due to its Stan backend.

NeuralProphet is a hybrid forecasting framework based on PyTorch and trained with standard deep learning methods, making it easy for developers to extend the framework.
Local context is introduced with auto-regression and covariate modules, which can be configured as classical linear regression or as Neural Networks. 
Otherwise, NeuralProphet retains the design philosophy of Prophet and provides the same basic model components.

Our results demonstrate that NeuralProphet produces interpretable forecast components of equivalent or superior quality to Prophet on a set of generated time series.
NeuralProphet outperforms Prophet on a diverse collection of real-world datasets. For short to medium-term forecasts, NeuralProphet improves forecast accuracy by 55 to 92 percent.

\end{abstract}


\end{frontmatter}

\section{Introduction}
\subsection{Background}
Time series data is prominent in most industrial sectors.
Though extensively studied in theory and applications such as economics, practical forecasting in industrial applications has not received widespread attention until recently. 

Since many companies have significantly improved their data collection to the extent where data availability now exceeds their data analytic capabilities, how to process and forecast industrial time series is quickly becoming an important topic. 
Often the person tasked with solving a particular issue is not a time series expert but rather a practitioner forecaster. 

Accurate time series forecasting is essential for decision-making processes, especially in infrastructure planning, budget allocation and supply chain management. Over-forecasts as well as under-forecasts can be both be costly. In applications, where forecasts inform business or operational decisions with potentially fatal consequences, it is often a requirement for the model to be explainable. A common approach is to break the forecast down into  components which are individually interpretable. This allows for the forecast to be reviewed by a domain expert who may make adjustments when appropriate, a procedure known as human-in-the-loop. 

\paragraph{Classical Time Series Methods}
The field of forecasting was traditionally dominated by statistical techniques. This was most notable on many of the early forecasting competitions such as NN3, NN5 and M3~\citep{m3, Crone2008}.
Classical models such as Auto-Regressive Integrated Moving Average (ARIMA) and Exponential Smoothing (ETS) have been well studied and provide interpretable components.  However, their restrictive assumptions and parametric nature limit their performance in real world applications. A skillful forecasting expert can transform data and combine algorithms to satisfy specific conditions for better performance. This requires deep domain knowledge in the application itself and in classical time series modelling. 

\paragraph{Machine Learning Methods}
Machine Learning (ML) based models were constantly performing poorly at the early forecasting competitions. Neural Networks (NN) were once even labeled as not being competitive for forecasting~\citep{hyndman2018blog}. They were further criticized in literature for their black-box nature, as in~\cite{Makridakis2018}. 
However, with the explosion in availability of large scale time series, NN-based data-driven techniques regained their popularity in forecasting. The amount of data available is no longer insufficient for ML and Deep Learning (DL) techniques to train well. 
Nevertheless, explainability of these models remains mostly an open research problem in the field of forecasting. 
Further, they often require substantial engineering efforts to preprocess data and fine-tune hyperparameters.

Consequently, most non-expert forecasting practitioners involved in different industries do not use the most accurate state-of-the-art models for their particular task. Rather, they are more interested in finding a reasonably accurate model, subject to explainability, scalability and minimal tuning. These are commonly seen as requirements for forecasting applications in industry.

Practitioners tend to opt for traditional statistical techniques which are user-friendly and interpretable in functionality, despite their poor forecasting performance. 
Hence, new methods need to be developed which can bridge this gap between classical time series modelling and ML-based methods.

\paragraph{Hybrid Methods}
A precursor to hybrid methods, Facebook Prophet~\citep{Taylor2017}, provides an interpretable model which scales to many forecasting applications. The forecasting framework provides full automation to novices and fine-tuning capabilities to domain experts.
Prophet remains one of the first forecasting packages which a data scientist will use, and also one of the few which they often continue to use as their skills grow. It has made classic time series forecasting approachable and useful to a wide demographic.

Yet, its limitations around key features, such as the lack of local context, and extensibility have presented challenges for users.
The lack of local context, which is essential for forecasting the near-term future, has restricted the usefulness of Prophet in industrial applications. Because Prophet was built on top of Stan~\citep{stan}, a probabilistic programming language, it is difficult to extend the original forecasting library.

\paragraph{Our Contribution}
Inspired by Prophet's impact on the forecasting community, it is our objective to continue the democratization of forecasting tooling, by making hybrid models as accessible as Prophet made classic models. 
As a step towards truly hybrid methods, we introduce NeuralProphet, a user-friendly and interpretable forecasting tool built on PyTorch~\citep{pytorch}.  NeuralProphet fuses the classic time series components introduced by the Prophet package with NN modules into a hybrid model enabling it to fit to non-linear dynamics. Two such NN components are the auto-regression and covariate modules. By using PyTorch as the backend, NeuralProphet can be updated with the latest innovation in deep learning.  
We provide a convenient tool for non-experts in forecasting by pre-selecting strong defaults and automating many of the modelling decisions. Advanced users may also find the package useful as they can incorporate domain knowledge with many customizaiton options. Overall, NeuralProphet abstracts a lot of forecasting domain knowledge and incorporates ML best practices so users can focus on the task at hand.

\paragraph{in this paper}

We introduce the NeuralProphet model and its components in detail and compare it to Prophet. 
The methods section describes the experimental setup and evaluation.
The synthetic data results quantify the accuracy of the decomposed interpretable forecast components.
The real-world benchmarks demonstrate the realistic out-of-the-box forecasting performance on a diverse set of univariate time series. 
Finally, we summarize the findings and provide suggestions on which model to use depending on the task at hand.

\section{The NeuralProphet Model}
\subsection{Model Components}

A core concept of the NeuralProphet model is its modular composability. The model is composed of modules which each contribute an additive component to the forecast. Most components can also be configured to be scaled by the trend for a multiplicative effect. Each module has its individual inputs and modelling processes. However, all modules must produce $h$  outputs, where $h$ defines the number of steps to be forecasted into the future at once. These are added up as the predicted values $\hat{y}_{t}, ..., \hat{y}_{t+h-1}$ for the time series future values $y_{t}, ..., y_{t+h-1}$. If the model is only time-dependent, an arbitrary number of forecasts can be produced. In the following descriptions, that special case will be treated mathematically equivalent to a one-step ahead forecast with $h=1$.

\begin{equation}
\label{eqn:model}
\hat{y}_t =  T(t) + S(t)  + E(t) + F(t) + A(t) + L(t)  
\end{equation}

where,
\begin{align*}
T(t) &= \text{Trend at time $t$} \\
S(t) &= \text{Seasonal effects at time $t$} \\
E(t) &= \text{Event and holiday effects at time $t$} \\
F(t) &= \text{Regression effects at time $t$ for future-known exogenous variables} \\
A(t) &= \text{Auto-regression effects at time $t$ based on past observations} \\
L(t) &= \text{Regression effects at time $t$ for lagged observations of exogenous variables} \\
\end{align*}

All model component modules can be individually configured and combined to compose the model. If all modules are switched off, only a static offset parameter is fitted as the trend component. By default, only the trend and seasonality modules are activated. The full model is summarized in equation~\ref{eqn:model}

In the following subsections we discuss each of the components in more detail.

\subsubsection{Trend}

A classic approach to modelling trend is to model it as the combination of an offset $m$ and a growth rate $k$. The trend effect at a time $t_1$ is given by multiplying the growth rate by the difference in time since the starting point $t_0$ on top of the offset $m$.

\begin{equation}
T(t_1) = T(t_0) + k \cdot \Delta_t = m + k \cdot(t_1 - t_0) 
\end{equation}

NeuralProphet models the trend component with this classic approach, but allows the growth rate to change at a number of locations. Thus, the trend is modelled as a continuous piece-wise linear series. This results in an interpretable, yet non-linear form of trend modelling. It is simple to interpret, as in a segment between two points, the trend effect is given by the steady growth rate multiplied by the difference in time. We can generalize the trend by defining a time-dependent growth rate $\delta(t)$ and a time-dependent offset $\rho(t)$.

\begin{equation*}
\begin{split}
T(t) &= \delta(t) \cdot t + \rho(t)
\end{split}
\label{eqn:trend_simple}
\end{equation*}

The piece-wise linear trend only varies the growth rate at a finite number of changepoints. A set $C$ of $n_{c}$ changepoints are defined at different times as $C=(c_1, c_2, ..., c_{n_c})$.  Between changepoints, the trend growth rate is kept constant. The first segment's growth rate and offset are given as $\delta_0$ and $\rho_0$ respectively.
Rate adjustments at each changepoint can be defined as a vector $\delta \in \mathbb{R}^{n_C}$, where $\delta_j$ is the rate change at the $j^{th}$ change-point.
The growth rate at time $t$ is determined by adding the initial growth rate $\delta_0$ with the summation of the rate adjustments at all the change-points up to time step t. Each growth rate change $\delta_j$ is a parameter to be fitted on the data. 
A corresponding vector of offset adjustments can be defined as $\rho \in \mathbb{R}^{n_c}$.
Similarly, the offset at time $t$ is given by the initial offset $\rho_0$ and the sum of the offset adjustments at each change-point up to time $t$. However, the offset for a changepoint $c_j$ is not an independent parameter but defined as $\rho_j = - c_j\delta_j$. This particular offset definition makes the piece-wise linear series continuous. 
Further, we introduce a binary vector $\Gamma(t) \in \mathbb{R}^{n_c}$ representing whether the time $t$ is past each changepoint. Thus, we can define a vectorized equation for the trend $T(t)$ at time $t$, which is given in equation~\ref{eqn:trend}:

\begin{equation}
\begin{split}
T(t) = (\delta_0 + \Gamma(t)^T\delta) \cdot t + (\rho_0 + \Gamma(t)^T \rho) 
\end{split}
\label{eqn:trend}
\end{equation}

where,
\begin{align*}
\delta &= (\delta_1, \delta_2, ..., \delta_{n_C})\\
\rho &= (\rho_1, \rho_2, ..., \rho_{n_C}) \\
\Gamma(t) &= (\Gamma_1(t), \Gamma_2(t), ...,  \Gamma_{n_c}(t)) 
\end{align*}
$$ \Gamma_j(t) = \begin{dcases*} 1, & if $ t \geq c_j$\\ 0, & otherwise \end{dcases*}$$

NeuralProphet provides a simple, semi-automatic mechanism for the selection of relevant change-points. Given the number $n_C$ of desired change-points, $n_C$ equidistant points along the series are selected as initial changepoints. Optionally, their growth change rate parameters can be regularized during model training. This is is similar to a fully automatic changepoint selection, as only the most relevant changepoints will be selected, if any. The user can also opt to manually define the specific times of a custom number of changepoints. To avoid overfitting on a small number of final points, the final trend segment (after the last changepoint) is set to a larger set of observations (default: 15 \% of training data). When making predictions into the unobserved future, the final growth rate is used to linearly extrapolate the trend.

\subsubsection{Seasonality}

Seasonality in NeuralProphet is handled by using Fourier terms~\citep{fourier} as was originally done in Prophet~\citep{Taylor2017}. 
In this technique, a number of Fourier terms are defined for each seasonality as in Equation \ref{eqn:seasonality_modelling}, where $k$ refers to the number of Fourier terms defined for the seasonality with periodicity $p$. Fourier terms are defined as sine, cosine pairs and allow to model multiple seasonalities as well as seasonalities having non-integer periodicities such as yearly seasonality with daily data ($p=365.25$) or with weekly data ($p=52.18$). In a multiple seasonality scenario, different values for $n$ can be defined for each periodicity.
\begin{equation}
\label{eqn:seasonality_modelling}
S_p(t) = \sum_{j=1}^{k}\left(a_j \cdot cos\left(\frac{2\pi jt}{p}\right) + b_j \cdot sin\left(\frac{2\pi jt}{p}\right)\right)
\end{equation}

Fourier terms are a great tool for modelling seasonality as they produce smooth functions which are simple to interpret and stable to fit to data. However, Fourier terms only model deterministic seasonal shapes which are assumed to be fixed through time. A higher number of Fourier terms allows the model to fit a more complex seasonal pattern. Too much flexibility may lead to overfitting or to random patterns between observations. Thus, each Fourier term corresponds to a frequency proportional to $\frac{j}{p}$, modelled by by a weighted combination of a sine and cosine transform. Every seasonality is associated with $2k$ number of coefficients. For time step $t$, the effect from all the seasonalities considered in the model can be indicated by $S(t)$ in below Equation \ref{eqn:all_seasonalities}, where $\mathbb{P}$ refers to the set of all the periodicities.

\begin{equation}
\label{eqn:all_seasonalities}
    S(t) = \sum_{p \in \mathbb{P}}{S^{\star}_p(t)}\\
\end{equation}

Both additive and multiplicative seasonal patterns are supported. 
Each seasonal periodicity $S_p^{\star}$ can individually be configured to be multiplicative, in which case the component is multiplied by the trend. 
\[
    S_p^{\star}(t) = \begin{dcases*}
S_p^{\dagger}(t) = T(t) \cdot S_p(t), & if $S_p$ is multiplicative \\
S_p(t), & otherwise\\
\end{dcases*}
\]

The framework automatically activates daily, weekly and or yearly seasonality depending on data frequency and length. Each of these three types of seasonal periodicities is activated if the data frequency is higher resolution than the respective periodicity, and if at least two full periods of data are available. As an example, if the data is of daily frequency, the model will enable yearly seasonality if the data spans two years or more. Weekly seasonality will also be added if two or more weeks of data are available. Daily seasonality will not be activated, as the daily frequency is not of higher resolution to allow for intra-day seasonality. Unless otherwise specified, the default number of Fourier terms per seasonality are: $k=6$ for $p=365.25$ yearly, $k=3$ for $p=7$ weekly, and $k=6$ for $p=1$ daily seasonality. 

\subsubsection{Auto-Regression}\label{sec:AR}
Auto-regression (AR) refers to the process of regressing a variable's future value against its past values, a critical part of many forecasting applications as can be seen in \cite{hyndman2014forecasting}. The number of past values included is usually referred to as the order $p$ of the AR$(p)$ model. Hereby, a coefficient  $\theta_i$ is fitted for each past value. Each coefficient $\theta_i$ controls the direction and magnitude of effect of a particular past value on the forecast.  In a classic AR process, an intercept $c$ and a white noise term $\epsilon_t$ are included.
$$ y_t=c+\sum_{i=1}^{i=p}{\theta_i \cdot y_{t-i}}+e_t $$

In many applications, we are interested to predict multiple steps into the future. We refer to the number of forecasts made at once as the forecast horizon $h$. A traditional AR model will only produce one prediction for a one-step-ahead forecast with $h=1$. If a multi-step ahead forecast of length $h>1$ is demanded, $h$ models will have to be fitted, one for each forecast step. The AR module in NP is based on a modified version of AR-Net in \cite{arnet}. AR-Net is able to produce all $h$ forecasts with one model, which can be of linear or non-linear nature. In any configuration, the $p$ last observations of the target variable $y_{t-1}, y_{t-2}, ..., y_{t-p}$, also referred to as lags, are the inputs to the module. The outputs are $h$ values corresponding to the AR-effect for each forecast step $A^t(t), A^t(t+1), ..., A^t(t+h-1)$. Hereby, $A^t(t+2)$ denotes the predicted AR-effect for forecast  $\hat{y}_{t+2}$ at $t+2$, three steps in the future, predicted at forecast origin $t$ with observed data up to and including $t-1$.

\begin{equation}
A^t(t), A^t(t+1), ..., A^t(t+h-1) = \text{AR-Net}(y_{t-1}, y_{t-2}, ..., y_{t-p}) 
\label{eqn:AR_ARNet}
\end{equation}

It is important to note, that each time we forecast at a specific origin, we obtain $h$ predictions. Thus, for a given point in time, we have up to $h$ different predictions, each originating from a different forecast made in the past. They differ from each other based on the data available to the model at the time of the forecast. As an example, when we use an $AR(5)$ model to forecast $h=3$ steps into the future, at a given time $t=3$, we will have 3 predicted values of the AR-effect for $\hat{y}_{t=3}$. These three effects $A^{t=1}(t=3), A^{t=2}(t=3), ..., A^{t=3}(t=3)$ are each an estimation of the true AR-effect $A(t=3)$, with different 'age'. The oldest estimate $A^{t=1}(t=3)$ is 3 steps old, while the most recent estimate $A^{t=3}(t=3)$ is one step old. With Auto-Regression, any estimate will be at least one step old, as we assume to never know the true value of the series at the current time $t$ but only the observed value at the last step $t-1$ and older.

\paragraph{AR order}
The most important parameter for this module is the number of past values to be regressed over, also known as the order $p$ of the AR$(p)$ model. This parameter should be chosen based on the approximate length of relevant context in past observations. In practice, it is hard to determine accurately and is commonly set to twice the innermost periodicity or twice the forecast horizon. Alternatively, a conservatively large order can be chosen when used in combination with regularization to obtain a sparse AR-model.

\paragraph{Linear AR}
The default AR-Net configuration does not contain hidden layers, and is functionally identical to a classic AR model. In practice, it is a single layer NN with $p$ inputs, $h$ outputs, no biases, and no activation function. The single layer weights each regress a particular lag onto a particular forecast step. Thus, each weight can be matched to a corresponding coefficient of a collection of $h$ classic AR($p$) models, making it simple to interpret the model.

$$ y = Wx$$

where,
\begin{align*}
x &= (y_{t-1}, y_{t-2}, ..., y_{t-p})\\
y &= (A^t(t), A^t(t+1), ..., A^t(t+h-1)) 
\end{align*}

We can denote the model as a vector-matrix multiplication for the predicted AR effects $y \in \mathbb{R}^{h}$, with lagged observations as input $x \in \mathbb{R}^{p}$, and with weight matrix $ W \in \mathbb{R}^{h \times p}$, where $W_{i,j}$ denotes the coefficent defining the linear impact of lag $y_j$ on AR-effect $A^t(t=i)$

\paragraph{Deep AR} 
AR-Net based AR module can model non-linear dynamics when hidden layers are configured. In this case, the module trains a fully connected Neural Network (NN) with the specified number of hidden layers and dimensions. 
The addition of hidden layers may lead to a better forecasting accuracy, however it a partial trade-off in interpretability. Instead of being able to directly quantify the contribution of a particular past observation to a particular prediction, we can only observe the relative importance of a given past observation on all predictions. This can be approximated by comparing the sums of the absolute weights of the first layer for each input position. 

The $p$ last observations of the time series are the inputs to the first layer. After each hidden layer, the logits are passed through an activation function, in our case, a rectified linear unit (ReLU). The final layer outputs $h$ logits, are not transformed by an activation function and have no bias. For $l$ hidden layers with a hidden layer dimension of size $d$, we have:

\begin{align*}
a_1 &= f_a( W_1 x + b_1 ) \\
a_i &= f_a( W_i a_{i-1} + b_i ) \text{ for } i \in [2, ..., l]\\
y &= W_{l+1} a_{l}   \\
\label{eqn:AR_deep}
\end{align*}

where,

\[
f_a(x) = ReLU(x) = \begin{dcases*}
    x, & $x \ge 0$ \\
    0, & $x < 0 $\\
\end{dcases*}
\]

Hereby, the layer biases all have the same dimensions $b \in \mathbb{R}^{d}$, while the layer weights are $W \in \mathbb{R}^{d \times d}$, except for the first  $W_1 \in \mathbb{R}^{d \times p}$ and last  $W_{l+1} \in \mathbb{R}^{h \times d}$ layer weights.

\paragraph{Sparse AR}
AR-Net demonstrated that the correct order can be approximated by setting the order to a slightly larger than expected value when regularization is used to sparsify the model weights. This allows for a more convenient model configuration while retaining interpretability. 

The regularization function proposed by the authors of AR-Net is given in equation~\ref{eqn:arnet-reg}. Though we make this regularization function available, by default, our implementation uses a different regularization function, which we found to work better for a wider range of data. We use the regularization function introduced in equation~\ref{eqn:regularization} with parameters  $\epsilon = 3$ and  $\alpha=1$, where $\theta $ is the vector of weights corresponding to the $p$ AR model coefficients:
\begin{equation}
\Lambda_{A}(\theta) = \Lambda(\theta, \epsilon=3, \alpha=1) =\frac{1}{p} \sum_{i=1}^p{ log(\frac{1}{3 \cdot e} + |\theta_i| ) + log(3) + 1 }
\end{equation}

where,
\[
\theta = \begin{dcases*}
    W \in \mathbb{R}^{p \cdot h}, & no hidden layers ($l=0$)\\
    W_1 \in \mathbb{R}^{p \cdot d}, & with hidden layers ($l \ge 1$)\\
\end{dcases*}
\]

As an example, when forecasting the next 24 steps of an hourly time series, the forecaster could choose to set the AR order to 100 with some regularization. After fitting, the AR module may then exhibit sparse coefficients with the most significant weights located on a few positions only. In this example, only positions 1, 2, 3, 24, 48, and 72 could have non-zero weights. These positions can be interpreted as the locations with an auto-correlation, and further, their weights can be examined to study how different combinations of past observation values affect the forecast.

For reference, the original regularization function proposed in AR-Net is given by:
\begin{equation}
    \label{eqn:arnet-reg}
    \Lambda_{AR-Net}(\theta, c_1, c_2) = \frac{1}{p} \sum_{i=1}^p{2 \cdot (1 + \exp(-c_1 \cdot |\theta_i|^{\frac{1}{c_2}})^{-1}} - 1
\end{equation}

where the authors suggest setting $ c_{1} \approx 3$ and $c_{2} \approx 3$.

\subsubsection{Lagged Regressors}
Lagged regressors are used to correlate other observed variables to our target time series. They are often referred to as covariates. Unlike future regressors, the future of lagged regressors is unknown to us. At the time $t$ of forecasting, we only have access to their observed, past values up to and including $t-1$. 

\begin{equation}
L(t) = \sum_{x \in \mathbb{X}}  L_x(x_{t-1}, x_{t-2}, ...,  x_{t-p}) 
\end{equation}

Given a set of  covariates $\mathbb{X} \in \mathbb{R}^{T \times n_l}$, we create a separate lagged regressor module for each of the $m$ covariates $x$ of lenght $T$. This allows to individually attribute the effect of each covariate on the predictions. Each lagged regressor module is functionally identical to the AR module, with the only difference being the inputs. Here, the $p$ last observations of the covariate $x$ are the inputs to the module (instead of the series $y$ itself as in AR).  The outputs are of identical form, each module producing $h$ additive components $L_x^t(t), L_x^t(t+1), ..., L_x^t(t+h-1)$ to the overall forecasts $\hat{y}_{t}, \hat{y}_{t+1},..., \hat{y}_{t+h-1}$.

\begin{equation}
L_x^t(t), L_x^t(t+1), ..., L_x^t(t+h-1) = \text{AR-Net}(x_{t-1}, x_{t-2}, ..., x_{t-p}) 
\label{eqn:L_ARNet}
\end{equation}

All considerations regarding the order, depth, sparsification, and interpretability of the AR-Net modules for each covariate are identical to the AR module, as discussed in section~\ref{sec:AR}, if  we substitute: 
\begin{align*}
x &= (x_{t-1}, x_{t-2}, ..., x_{t-p})\\
y &= (L_x^t(t), L_x^t(t+1), ..., L_x^t(t+h-1)) 
\end{align*}

\subsubsection{Future Regressors}

To model future regressors, both past and future values of these regressors have to be known. Given the set of future regressors as $\mathbb{F} \in \mathbb{R}^{T \times n_f}$, where $n_f$ is the number of regressors, the effect from all future regressors at time step $t$ can be denoted by $F(t)$ as in Equation \ref{eqn:future_regressor_modelling}, where $d_f$ stands for the coefficient of the model for future regressor $f \in \mathbb{V}$. By default, future regressors have an additive effect, which can be configured to be multiplicative instead.

\begin{equation}
\label{eqn:future_regressor_modelling}
F(t) = \sum_{f \in \mathbb{F}} F_f^{\star}(t)\\
\end{equation}

where,
$$F_f(t) = d_f f(t)$$
\[
F_f^{\star}(t) = \begin{dcases*}
F_f^{\dagger}(t) = T(t) \cdot F_f(t), & if $f$ is multiplicative \\
F_f(t), & otherwise\\
\end{dcases*}
\]



\subsubsection{Events and Holidays}\label{subseq:events}

Effects from special events or holidays may occur sporadically. Such events are modelled analogous to future regressors, with each event $e$ as a binary variable $e \in [0, 1]$, signaling whether the event occurs on the particular day or not. For a set of events $\mathbb{E} \in \mathbb{R}^{T\times n_e}$ with $n_e$ number of events and the length of the series $T$, the effect from all events at time step $t$ can be denoted by $E(t)$ in Equation \ref{eqn:events}, where $z_e$ denotes the coefficient of the model corresponding to the event $e \in \mathbb{E}$.

\begin{equation}
\label{eqn:events}
E(t) = \sum_{e \in \mathbb{E}} E_e^{\star}(t)\\
\end{equation}

where,
$$E_e(t) = z_e e(t)$$
\[
E_e^{\star}(t) = \begin{dcases*}
E_e^{\dagger}(t) = T(t) \cdot E_e(t), & if $e$ is multiplicative \\
E_e(t), & otherwise\\
\end{dcases*}
\]

NeuralProphet allows the modelling of two types of events; 1) user defined 2) country specific holidays. Given a country name, its national holidays are automatically retrieved and added to the set of events $\mathbb{E}$. Similar to seasonal effects, events can also be specified as either additive or multiplicative.

 Additionally, for a given event at time $t_e$, a window $[t_e-i, t_e+j]$ of $i+j$ days can be configured to be considered as special events of their own. For example, by setting a window of $[-1, 0]$ for Christmas day, will allow the day before Christmas to have its own effect on the forecast. Hereby, a new variable is created for each day within the window around event and added to the set of events $\mathbb{E}$. 

\subsection{Preprocessing}
\subsubsection{Missing Data}
Missing data is less of an issue when working with non-lagged input variables, as corresponding timesteps can simply be dropped. In doing so, one data sample will be lost per missing entry. With Auto-regression or lagged regressor modules in use, a missing data point would lead to $h+p$ data samples being dropped due to $h$ missing forecast targets and $p$ missing lag values. For example, a single missing point leads to the loss of 13 samples for a model forecasting $h=3$ steps ahead with AR($p=10$).
Thus, we implemented a data imputation mechanism to avoid excessive data loss when working with incomplete data. The imputation mechanism follows the following heuristics:

\paragraph{Data Imputation} If not specified or missing, events are assumed to not be happening. Missing Events are filled in with zeros, indicating their absence.
All other real-valued regressor variables, including the time series itself, if autoregression is enabled, are imputed in a three step procedure. 
\paragraph{First} The  missing values are approximated by a bi-directional linear interpolation. Hereby, the last and the first known value before and after the missing values are used as anchor points for the interpolation. This is done for up to 5 missing values in each direction. If there are more than 10 missing values, they will remain \textit{NAN} after this step. The amount of missing values to interpolate is user-settable.
\paragraph{Second} The remaining missing values are imputed with a centred rolling average. The rolling average is computed over a window of 30, and fills at most 20 consecutive missing values. The amount of missing values to fill with a rolling average is user-settable.
\paragraph{Third} If there are more than 30 consecutive missing values, the imputation algorithm aborts and instead drops all the missing datapoints.

\subsubsection{Data Normalization}
The type of normalization to apply to the time series can be set by the user. The available options are described in table~\ref{tab:data-normalizatoin}.
If not specified, or set to 'auto', the default option 'soft' is used, unless the time series values are binary, in which case 'minmax' is applied.

\begin{table}[htbp]
\centering
	\begin{tabular}{ll}
	\hline
	Name & Normalization Procedure\\
	\hline
	'auto' & 'minmax' if binary, else 'soft' \\
    'off' &  bypasses data normalization \\
    'minmax' & scales the minimum value to 0.0 and the maximum value to 1.0 \\
    'standardize' & zero-centers and divides by the standard deviation \\
    'soft' & scales the minimum value to 0.0 and the 95th quantile to 1.0 \\
    'soft1' & scales the minimum value to 0.1 and the 90th quantile to 0.9 \\
	\hline
	\end{tabular}
	\caption{The available data normalization options.}
	\label{tab:data-normalizatoin}
\end{table}

\subsubsection{Tabularization}
We tabularize the time series data to create a pseudo independent and identically distributed dataset, as is required for SGD based training.
Hereby, a data sample is created for each available time-stamp of the target time series 'y'. A sample includes a normalized time-stamp, and all inputs required by each configured module.

The modules Auto-regression and lagged covariates hereby extract the $p$ values before a given time-stamp from their respective time series and store them in a vector of size $p$ serving as input to the model.
If forecasting multiple steps ahead with $h>1$, the forecast targets also are stored in a $h$-sized vector for each time-stamp.

This is not a memory-efficient approach to prepare the dataset to training. Hoever, we chose this procedure due to its simplicity and for its compute-efficiency when training.

Analogous, the Fourier terms are prepared by computing the $2k$ different sine and cosine transforms of the standardized time component, for each of the configured seasonal periodicities.

\subsection{Training}

One of the fundamental differences between Prophet and NeuralProphet is the fitting procedure. Prophet utilizes L-BFGS~\citep{lbfsg}, implemented in Stan~\citep{stan} for fitting model parameters to the data. 
NeuralProphet retools Prophet, from the bottom up, replacing Stan with PyTorch, which is both flexible and easy to use. 

NeuralProphet relies on a modern version of  mini-batch stochastic gradient descent (SGD). This fitting procedure is compatible with the vast majority of deep learning methods, can be scaled reliably to large datasets, and can accommodate more complex model components. 
Any model component which is trainable by SGD can be included as a module. This makes it easy for developers to extend the framework with new features, and to adopt new research.

\subsubsection{Loss Function}
The default loss function is the Huber loss, also known as smooth L1-loss, which can be seen in equation~\ref{eqn:huber}. Below a given threshhold $\beta$, it is equivalent to mean squared error (MSE).  For values larger than $\beta$ it is equivalent to the mean absolute error (MAE). Compared to a pure MSE loss, it is less sensitive to outliers and may help prevent exploding gradients~\cite{fastrcnn}. We chose $\beta=1$ as threshold. 
Users can however choose to alternatively opt for MSE or MAE loss, or any other available or self-implemented function matching the PyTorch loss function format as their loss function.

\begin{equation}
\label{eqn:huber}
    L_{huber}(y, \hat{y}) = \begin{dcases*}
    \frac{1}{2 \beta} (y - \hat{y})^2, & for $| y - \hat{y} | < \beta$ \\
    |y - \hat{y}| - \frac{\beta}{2} , & otherwise
    \end{dcases*}
\end{equation}

\subsubsection{Regularization}

We use a scaled and shifted log-transform of the absolute weight values as general regularization function. For a module with model weights $\theta$ it is parametrized as follows:

\begin{equation}
    \label{eqn:regularization}
    \Lambda(\theta, \epsilon, \alpha) = \frac{1}{n} \sum_{i=1}^n{ log(\frac{1}{\epsilon \cdot e} + \alpha \cdot |\theta_i| ) + log(\epsilon) + 1 }
\end{equation}

where $\epsilon \in (0, \infty) $ sets the inverse of the offset and $\alpha \in (0, \infty) $ sets the scaling of the log-transform.
Higher values of $\epsilon$ lead to a steeper curve for weights close to zero. The parameter $\epsilon$ could be described as a control of the \textit{eagerness} to sparsify weights. Higher values of $\alpha$ lead to a flatter curve for weights of larger magnitude. The parameter $\alpha$ can be interpreted as a control of the \textit{acceptance} of large weights. As default values, we suggest $\epsilon = 1$ and  $\alpha=1$ for most applications.
$$ \Lambda(\theta, \epsilon=1, \alpha=1) = \frac{1}{n} \sum_{i=1}^n{ log(\frac{1}{e} + |\theta_i| ) + 1 } $$

The regularization is applied in the per-module configured strength and added to the loss function, to be back-propagated. The regularization only commences after a specified percentage of training, by default, after $50\%$, and is thereafter linearly increased from zero to its full configured strength at the end of training.

\subsubsection{Optimizer}
Analogous to the loss function, any optimizer matching the signature of a PyTorch optimizer can be configured.
As a reliable default, the AdamW optimizer~\citep{adamw} is used, unless otherwise specified. In our testing we observed AdamW to fit most reliably, with a slight tendency to overfit. We initialize AdamW with the configured learning rate, and set $\beta=(0.9, 0.999)$, $ \epsilon=1e-08$, and a weight decay of $1e-04$.

As a fallback option, we also provide a classic SGD optimizer with momentum set to 0.9, and weight decay set to $1e-4$. In our testing, we found SGD to lead to better validation performance, but at the expense of occasional divergence. If a user is willing to fine-tune training related hyperparameters, SGD poses an attractive alternative to AdamW.

\subsubsection{Learning Rate}
As we do not want to require users to be experts in machine learning, we make the learning rate, an essential hyperparameter for any NN, optional. We achieve this by relying on a simple but reasonably effective approach to estimate a learning rate, introduced as a learning rate range test in~\cite{lrtest}.

A learning rate range test is executed for $100 + log_{10}(10 + T) * 50)$ iterations, starting at $\eta = 1e-7$, ending at $\eta = 1e+2$ with the configured batch-size. After each iteration, the learning rate is increased exponentially until the final learning rate is reached in the last iteration.  Excluding the first 10 and last 5 iterations, the steepest learning rate is defined as the resulting learning rate. The steepest learning rate is found by selecting the learning rate at the position which maximizes the negative gradient of the losses.

In order to increase reliability, we perform the the test three times and take the log10-mean of the three runs $\eta_1, \eta_2, \eta_3$ as the selected learning rate $\eta$:

$$\eta^{\star} = \frac{1}{3} (log_{10}(\eta_1) + log_{10}(\eta_2) + log_{10}(\eta_3)) $$
$$\eta = 10^{\eta^{\star}} $$

\subsubsection{Batch Size}
The batch size $B$ is an optional parameter. If not user specified, the following heuristic will determine the batch size based on the length $T$ of the dataset:

\begin{align*}
B^{\star} &= 2^{2 + \lfloor \log_{10}(T) \rfloor}  \\
B &= \min(T, \max(16, \min(256, B^{\star} )))
\end{align*}

\subsubsection{Training Epochs}
The number of training epochs $N_{epoch}$ is an optional parameter. If not user specified, the following heuristic will determine the number of epochs :

\begin{align*}
N_{epoch}^{\star} &= \frac{1000 \cdot 2^{\frac{5}{2}  \cdot \log_{10}(T)}}{T} \\
N_{epoch} &= \min(500, \max(50, \lfloor N_{epoch}^{\star} \rfloor ))
\end{align*}

\subsubsection{Scheduler}

As all training related hyperparameters are automatically approximated, none of them are optimal, and may thus potentially lead to training issues.
Nonetheless, a user expects the model to be reasonably fitted to the data without issues. 
In order to meet these expectations despite suboptimal hyperparameters, we rely on the training schedule called '1cycle' policy, which allows for 'superconvergence' of NN training, according to~\cite{superconvergence}. 
Hereby, the initial learning rate $\frac{\eta}{100}$ is gradually increased up to the peak learning rate $\eta$, reached at $30\%$ of training. Thereafter, the learning rate is annealed along a cosine curve down to $\frac{\eta}{5000}$ at the end of training.

\subsection{Postprocessing}
At the end of training or predicting, the normalized values on which the model operates are transformed back into the initial distribution. This, alongside with the many heuristically or procedurally set optional hyperparameters abstract a lot of the machine learning specific and time series specific decision making, making NeuralProphet approachable to forecasting practitioners.

\subsubsection{Metrics}
A few metrics are recorded by default: The configured loss function, RMSE, and MAE. Further, any user defined metrics will also be recorded over the training and validation runs.

\begin{equation}
    RMSE = \sqrt{\frac{1}{T}\sum_{i=1}^{T}(y_i - \hat{y}_i)^2}
\end{equation}

\begin{equation}
    MAE = \frac{1}{T}\sum_{i=1}^{T} | y_i - \hat{y}_i |
\end{equation}
\begin{align*} 
\textnormal{where} & \\
& T = \textnormal{size of data} \\
\end{align*}

\subsubsection{Forecast Presentation}
When used to predict, the model returns a dataframe with a column for each forecast component. These are the additive (or multiplicative) contributions of each component. They are individually displayed for each forecasted step ahead.
Each of the components is referring to its row's datestamp-located target and ordered by the age of the foregast. e.g. 'yhat3' refers to the predicted $y$-value for the current location, based on data available three steps ago. Similarly, 'ar2' refers to the AR-effect predicted for the current location, based on data available two steps ago.

There are different plotting utilities available for visualizing the forecast itself, the forecast decomposition, the model parameters, and for specific forecast horizons of interest.

\section{Experimental Setup}

The experiments in this work serve to contrast the capabilities of NeuralProphet with Prophet. First, we compare the accuracy of the interpretable forecast components of Prophet and NeuralProphet on a set of synthetic datasets. Second, we compare the performance of the two frameworks on a number of real-world datasets. In the following, these experiments are described in detail.

\subsection{Decomposition Demonstration on Synthetic Data}
These experiments contrast the ability of Prophet and NeuralProphet to decompose a time series into individual components. 
Since we are interested in the decomposition accuracy of the different modelling components, we must know their ground truth. Due to a lack of suitable real-world datasets with known underlying components, we create synthetic datasets as the sum of a selection of generated time series with added noise. This allows us to compare the estimated decomposed components of both models to the original components. 

\subsubsection{Synthetic Dataset Composition}
We generate the individual component time series to be identical to the type of components that Prophet and NeuralProphet claim to model. Thus, ideally, both models should be able to perfectly decompose the time series into its components, with some error due to additive noise included in the synthetic datasets.

\begin{table}[htbp]
\centering
	\begin{tabular}{cccccccc}
	\hline
	Experiment & Trend & Seasonality  & Events & Future reg. & AR & Lagged reg.\\
	\hline
	S-TS & \plusmark &  \plusmark && &&\\
	S-EF &  &  &\plusmark & \plusmark &&\\
	S-TSEF & \plusmark & \plusmark &\plusmark & \plusmark &&\\
	S-mTSEF & \plusmark & \plusmark$^\text{\smark}$ & \plusmark$^\text{\smark}$ & \plusmark$^\text{\smark}$ &&\\
	S-AL &  & &  & & \plusmark & \plusmark \\
	S-TSAL & \plusmark & \plusmark  & & & \plusmark & \plusmark\\
	S-TSEFAL & \plusmark & \plusmark  & \plusmark & \plusmark & \plusmark & \plusmark\\
	\hline
	\end{tabular}
	\caption{Synthetic components included in the experimental scenarios. \smark~ indicates that the respective component is multiplied by the trend.}
	\label{tab:S_experimental_scenarios}
\end{table}

In Table \ref{tab:S_experimental_scenarios} we display which components are included in the synthetic dataset of each experimental scenario. The comparison between the two models focuses on the accuracy of individual decomposed components. We also measure the overall accuracy and the compute time. Hereby, the emphasis is on the accuracy of the individual components as this quantifies the decomposition accuracy of the models.

As many real-world datasets exhibit auto-regressive properties or depend on lagged regressors, we also include three experiments with lagged components.  However, only NeuralProphet claims to be able to model Auto-Regression and Lagged regressor components. In experiments including such lagged components, Prophet's prediction for these components will be marked as zero.

Due to the freedom to choose offsets in their additive components, the models may find interpretable components which match the dynamics of the underlying components, but shifted by an offset. Thus, before computing any component-wise metrics, we zero-center all components.

\subsubsection{Generated Component Time Series} 
Each experiment's dataset consists of 5 independently generated time series of daily resolution with a length of 6000 samples each. Each time series is composed as the aggregate of multiple generated time series, each representing a particular component type. A component time series is scaled to range $[0, 1]$ before being aggregated with the other components. The aggregate time series is finally re-scaled to range $[0, 1]$. To denote the building blocks of the synthetic experiments we use the following notation where each component is represented by a letter:

\paragraph{\textbf{T}: Trend} A piece-wise linear trend with one random change-point. Trend increases linearly before the change-point, and decreases linearly after the change-point.

\paragraph{\textbf{S}: Seasonality} The generated seasonal component time series consists of Fourier terms combined with random weights. The components are generated based on equation \ref{eqn:seasonality_modelling}, with $a_j$ and $b_j$ drawn from a random uniform distribution such that $a_j \in [0,1)$ and $b_j \in [0,1)$ and $k=5$. For the experiments we create two independent seasonal components: monthly $S_{30}(t)$ and yearly $S_{365}(t)$. Both are added to the aggregate series and both are individually evaluated in their decomposition accuracy.

\paragraph{\textbf{E}: Events} The event component is a binary series with 25 occurrences of an event at random locations. Analogous to equation~\ref{eqn:events}, we have $E = (e^1) \in \mathbb{R}^{T\times1}$, with $e^1 = (e^1_1, ..., e^1_T) \in \mathbb{R}^{T}$ where $e^1_i = 1$ if an event of type $e^1$ occurs at timestamp $t_i$ and 0 otherwise. The initial effect strength $z_e=1$ will effectively become $z_e \geq \frac{1}{N_{components}}$ after scaling the main series.

\paragraph{\textbf{F}: Future Regressor} The future regressor component is generated as an AR(3) process with coefficients $\phi_1=0.2, \phi_2=0.3, \phi_3=-0.5$ and additive white noise. After scaling the series to a range of $[0,1]$, the regressor is added to the aggregate series with a weight of $d_f=1$, which will be $d_f \geq \frac{1}{N_{components}}$ after scaling the main series.

\paragraph{\textbf{A}: Auto-Regression} The Auto-regressive component is drawn from an AR(2) process with coefficients $\phi_1=0.3, \phi_2=0.3$, and additive white noise. This series is generated independent of the the other components. This is not identical to how NeuralProphet models the auto-regressive behavior. NeuralProphet does not decompose the inputs to auto-regression module, but rather uses the raw time series values as inputs. 

\paragraph{\textbf{L}: Lagged Regressor} First, an independent time series $x$ is generated with an AR(2) process, analogous to future regressors. Next, we create the lagged regressor effect series $L(t) \in \mathbb{R}^T$, where each effect depends on a weighted combination of the last three observations from series $x$ :
$$L(t) = c_1 \cdot x_{t-1} + c_2 \cdot x_{t-2} + c_3 \cdot x_{t-3}$$
where $c_1, c_2, ..., c_{lag}$ are weights from a random uniform distribution on $(0, 1]$.
Finally, we utilize the series $L(t)$ as the lagged regressor component.

\paragraph{\textbf{m}: multiplicative mode} In multiplicative mode, we multiply each component element-wise by the trend. For example, in S-mTSEF: 
\begin{align*}
    y_t &= T(t) + E^{\dagger}(t) + S^{\dagger}(t) +F^{\dagger}(t))\\
    &= T(t)\cdot(1 + E(t) + S(t) +F(t))
\end{align*}

\subsection{Main Benchmark on Real-World Datasets}

We evaluate the forecasting accuracy of Prophet and NeuralProphet on a diverse collection of datasets covering many univariate forecasting tasks. The datasets span data lengths from one hundred to half a million samples, while data record intervals span minutely to monthly frequencies. This wide spectrum of univariate time series is in stark contrast to most benchmarking efforts focused on NN-based methods. The results represent realistic out-of-the-box performance across applications such as energy demand, tourism, environmental factors, and retail sales among others. 

In this section, we present the selected datasets and the evaluation procedure in more detail.

\begin{figure}[htbp]
\centering
\begin{subfigure}[t]{0.33\textwidth}
  \centering
    \includegraphics[width=1.1\textwidth]{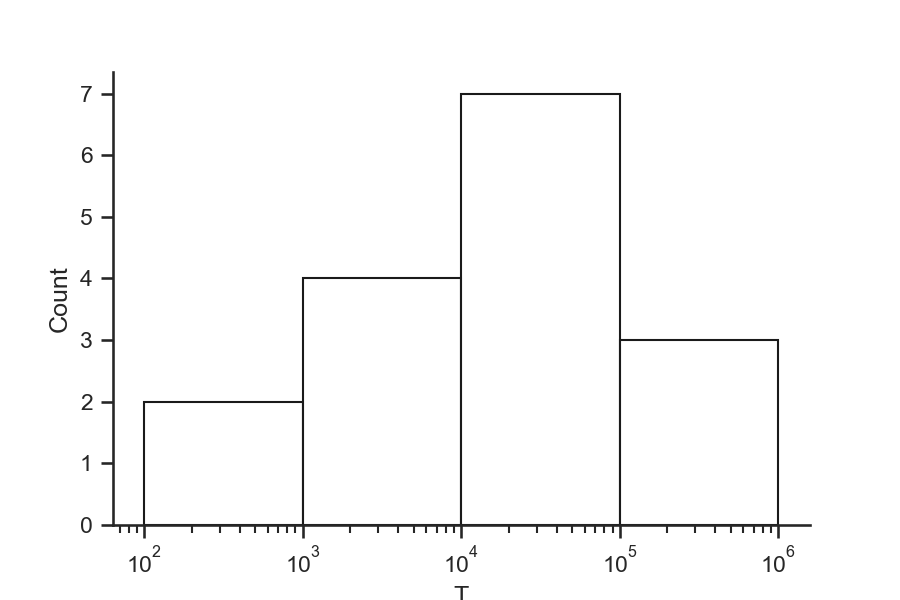}
    \caption{Histogram of lengths.}
    \label{fig:data-len}
\end{subfigure}%
\begin{subfigure}[t]{.33\textwidth}
  \centering
  \includegraphics[width=1.1\linewidth]{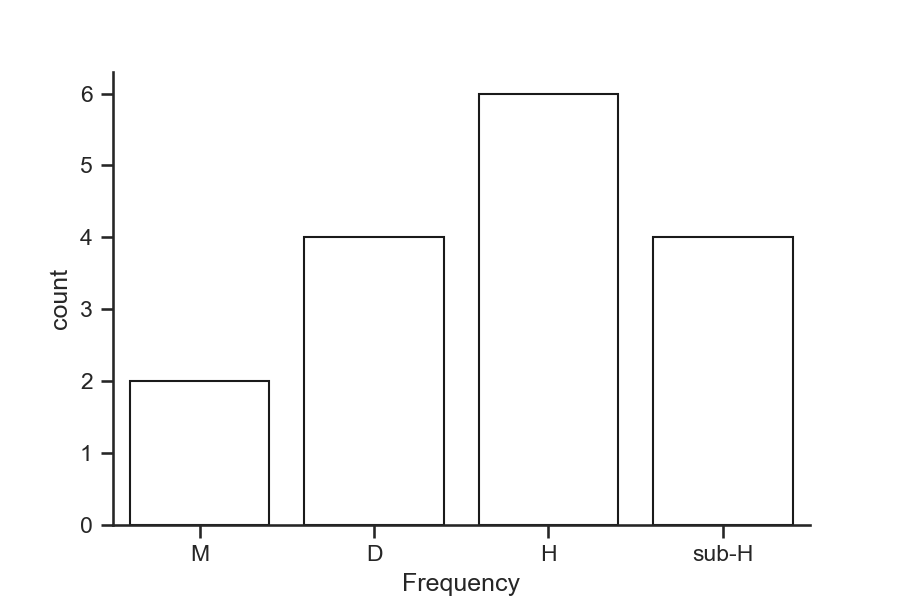}
  \caption{Counts of sampling frequencies.}
  \label{fig:data-freq}
\end{subfigure}
\begin{subfigure}[t]{.33\textwidth}
  \centering
  \includegraphics[width=1.1\linewidth]{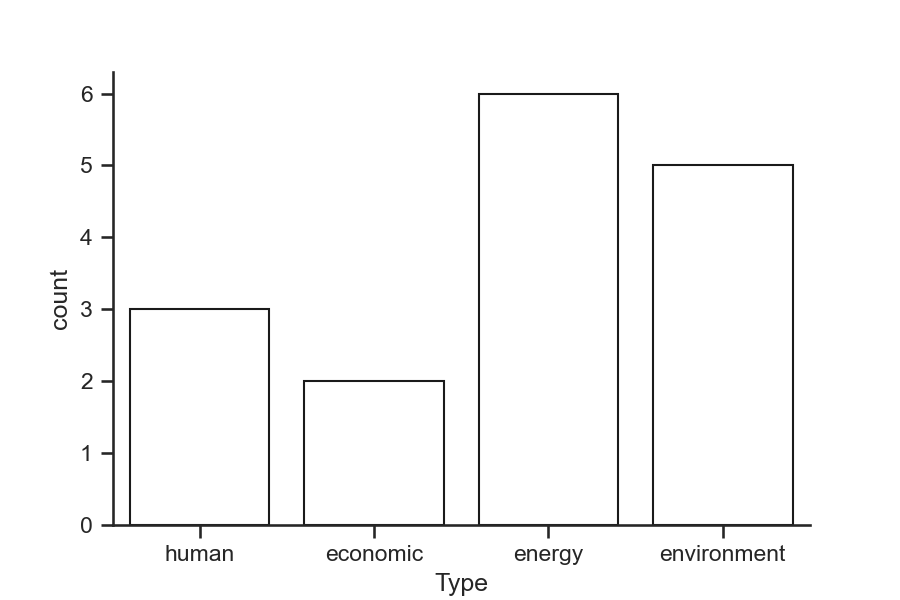}
  \caption{Counts of data types.}
  \label{fig:data-type}
\end{subfigure}
\caption{Histograms of different characteristics of the datasets.}
\label{fig:data-stats}
\end{figure}

\subsubsection{Selected Datasets}

The datasets are selected to represent a wide range of univariate datasets. Figure~\ref{fig:data-len} shows that the lengths of datasets range from one hundred to one million samples, with most datasets having around 10,000 samples. The data observation frequencies are distributed from monthly to minutely sampling frequencies. Figure~\ref{fig:data-freq} bins minutely to half-hourly frequencies together and displays the count of each type of frequency, with hourly frequencies being the most common. 

Our collection of testing datasets covers a wide variety of subjects, including environmental, energy-related, social, and economic data, with environmental and energy-related datasets being the most common (Figure~\ref{fig:data-type}).

\begin{figure}[htbp]
\centering
\begin{subfigure}[t]{.45\textwidth}
  \centering
  \includegraphics[width=1\linewidth]{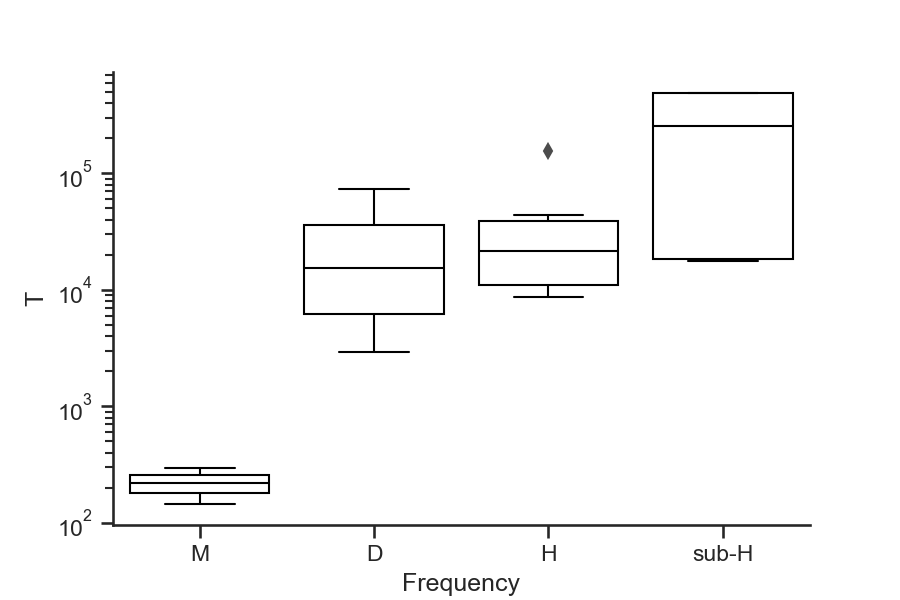}
  \caption{Boxplot of length for data frequencies.}
  \label{fig:data-freq-len}
\end{subfigure}
\begin{subfigure}[t]{.45\textwidth}
  \centering
  \includegraphics[width=1\linewidth]{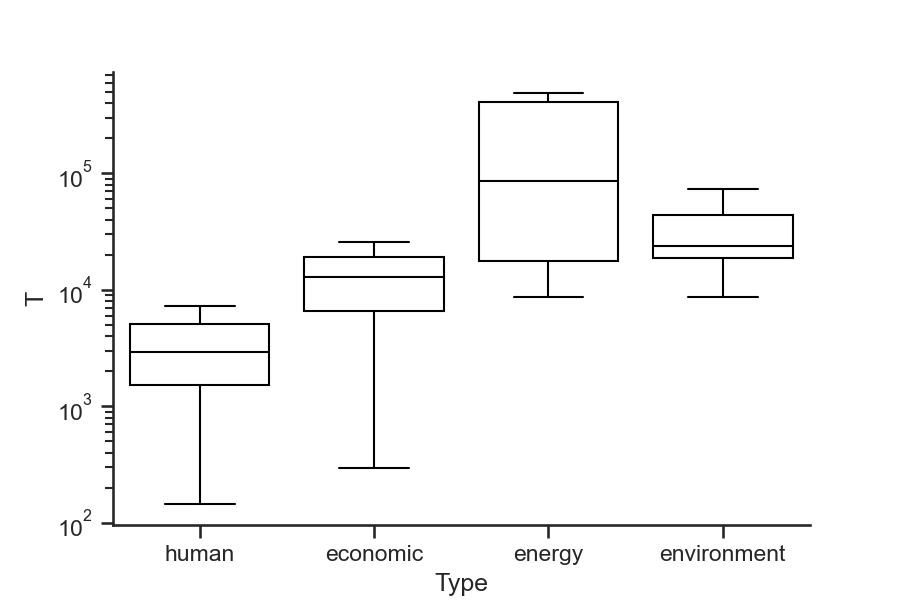}
  \caption{Boxplot of length for data types.}
  \label{fig:data-type-len}
\end{subfigure}
\caption{Distribution of data length for different data frequencies and data types.}
\label{fig:data-stats-len}
\end{figure}

Quantitative criteria for inclusion in our collection of testing datasets were: Univariate time series of minutely to daily sampling frequency, length of 1000 samples or more, and permissive license. We also include the four time series which Prophet uses as demonstration time series. Two of those datasets do not meet the inclusion criteria, as they are of monthly frequency and much less than 1000 samples long. They are included nonetheless, to make sure the datasets are not biased towards the proposed model NeuralProphet, but rather towards the comparison model Prophet.
Further details on each dataset can be found in Table~\ref{tab:datasets}  (Appendix A).

\paragraph{Sources} Datasets are taken from various sources including the UCI Machine Learning repository~\citep{Dua2019}, Monash time series repository~\citep{monashrepo}, and Prophet Github repository~\citep{Taylor2017}. 
Additionally, we collected and processed data from public energy and government webpages, including \cite{solarsf}, \cite{loadercot}, \cite{loadrte}, and \cite{loadhospital}.

\paragraph{Preprocessing} Most of the selected datasets do not contain missing values, but for those that do, we impute the values using linear interpolation unless otherwise described. However, for datasets which are mostly zero or near-zero-valued, such as data covering solar, parking, wind, and air quality subjects, we fill missing values with zeros.
Missing time-stamps are also regarded as missing data. Further, we manually preprocess each dataset and fix minor misalignments, such as time-stamps of 01:59 instead of 02:00 on data of hourly sampling frequency.

\subsubsection{Benchmark Models}
\label{sec:benchmarks}

In this work, we compare the performance of time series forecasting packages which support full model automation, offer interpretability and scale to large datasets, without requiring domain knowledge in machine learning or time series. Prophet was the first such package, which we benchmark against its only fully featured successor, NeuralProphet. 

\paragraph{Hyperparameters} Given that we aim to demonstrate the real world performance a non-expert forecasting practitioner can reasonably expect, we do not manually tune any hyperparameters. We only manually set hyperparameters in the synthetic benchmarking section if no automatic setting is available, or if needed to make models comparable.  

We mention any manually set parameters explicitly when discussing results. Occurrences of manual model configuration, but not hyperparameter tuning, include:  When studying the effect of adding auto-regression to the model, we set different numbers of lags $p$, which are all shown in the results. For multi-step ahead forecasts, we further need to set the forecast horizon $h$. In a reverse ablation study we further examine the impact of adding hidden layers to the model, hereby we also set the learning rate to the same value of $0.001$ for all of the model configurations and all datasets.

These ablation studies of number of lags, number of forecasts and neural network configurations will also serve to quantify the sensitivity of NeuralProphet to parameter choices.

\subsection{Forecast Evaluation}

In the following, we describe the forecasting task and how the forecast performance is evaluated.

\subsubsection{Forecast Horizon}
In this study, we are interested in univariate time series forecasting covering typical forecasting tasks.
Given a desired number of steps to forecast into the future, we predict the next few consecutive values of the series, starting from the present timestamp. 
The number of next consecutive values to be forecasted from the last observed value is also called forecast horizon. 
If auto-regression is configured, the model will regress over the last few observations of the series itself. 
We treat each given forecasting horizon as a separate forecasting task.

A forecasting task with a horizon of ten targets involves predicting the next ten targets starting at the current timestamp $t$, i.e. obtaining a series $\{\hat{y}_t, \hat{y}_{t+1}, ..., \hat{y}_{t+9}\}$ containing the predictions over the forecast horizon.
In our empirical evaluation over real-world datasets, we evaluate the forecast horizons of $[\infty, 1, 3, 15, 60]$. Hereby, a horizon of $\infty$ refers to forecasting the entire test set at once, as is possible with model based on time-features only, such as Prophet and default NeuralProphet.

\subsubsection{Expanding Origin Backtest}
Evaluation procedures in time series forecasting are diverse in nature. Naming conventions have only been establish for evaluations of classical time series models (\cite{evaluation}). Conventions for evaluating machine learning based time series forecasting models are currently being established. We utilize a method based on a fusion of cross-validation and backtesting, which originate from machine learning and trading algorithm evaluation. This method, called expanding origin backtest, has recently been adopting by researchers and companies trying to evaluate machine learning based models on time series data (\cite{uberbacktest, uberbacktest2}). 

The procedure consists of an outer and an inner loop. The term expanding origin backtest mostly refers to the outer loop, but is often used to describe the whole procedure. The inner loop is usually based on a variation of the classical evaluation procedures. In our case, the inner loop is a rolling origin forecast evaluation with non-constant holdout size (\citep{adambacktest}) with updating but without refitting of the model. In classical time series literature, refitting is referred to as reparametrization, while updating the model refers to updating the inputs to the model to produce the next forecast without refitting of model parameters. This means that a model will be re-fitted only $k$ times per evaluation procedure.

A more verbose name for our forecast evaluation procedure could be: K-fold expanding origin backtest with an inner rolling origin multi-step forecast evaluation with updating but without refitting of the model.

\begin{figure}[htbp]
    \centering
    \includegraphics[width=0.6\textwidth]{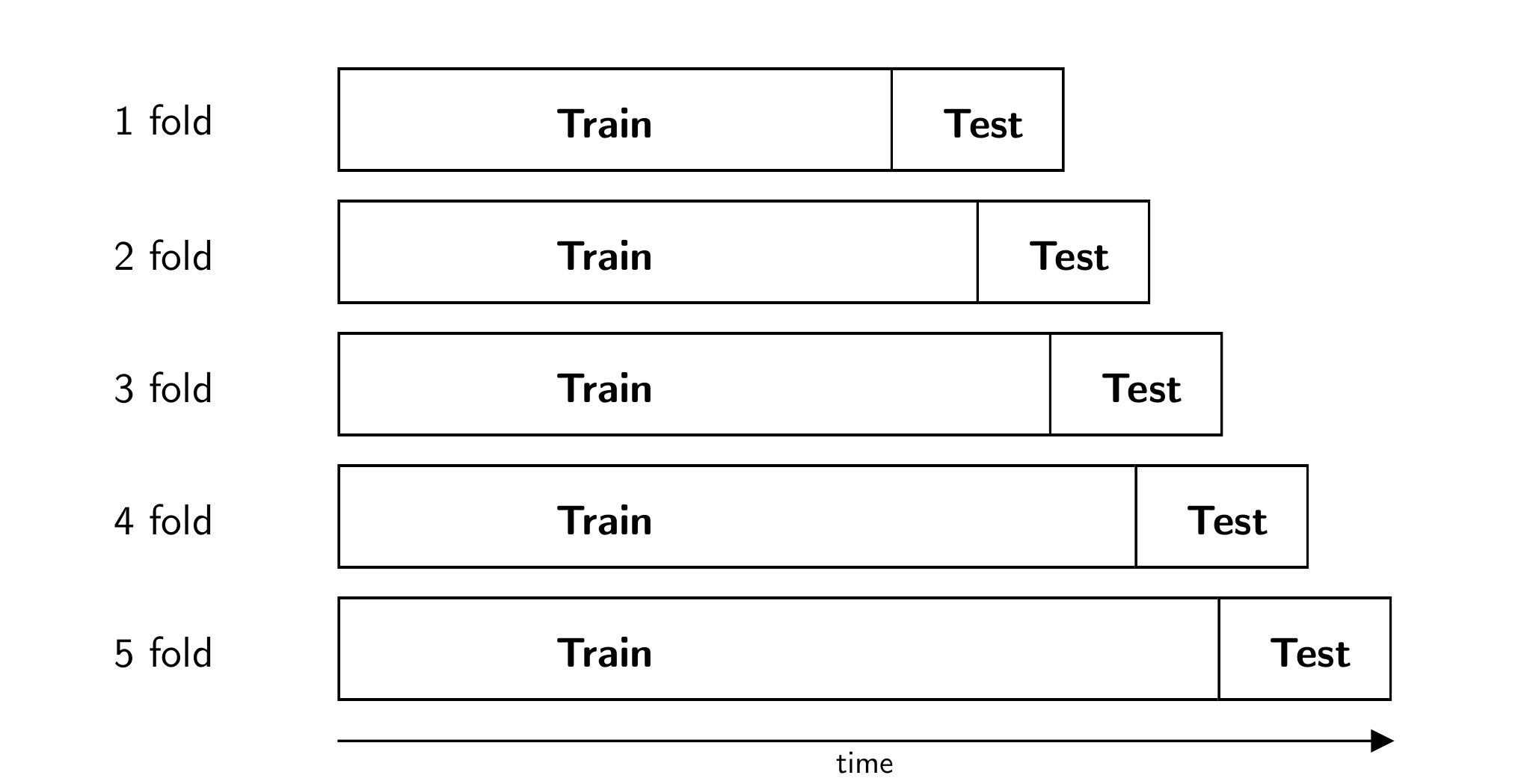}
    \caption{Expanding Origin Backtest with 5 folds. Test sets contain 10\% of the time series each.}
    \label{fig:backtest}
\end{figure}

\paragraph{Outer Loop: K-Fold Expanding Origin Backtest With Constant Holdout Size}
Analogous to $k$-fold cross-validation, the overall dataset is replicated into $k$ folds, each containing training and testing data, covering different parts of the overall dataset. The model to be evaluated is fitted separately on each fold's training data and evaluated on the same folds testing data. This signifies that the model 

Unlike $k$-fold cross-validation, the folds are not randomized. Each fold has a training set with values strictly occurring before the same fold's test set, a visualization can be seen in Fig.~\ref{fig:backtest}.

We perform a 5-fold expanding origin backtest with test sets of 10 percent and a forward rolling of the origin on 5 percent per fold. This means that each test fold will share 50 percent of their data with the previous fold, and that each train fold will grow by 5 percent, while the size of the test fold remains constant. Hereby, the first fold will use the first 70 percent of data for training and the next 10 percent for test. The fifth fold will utilize 90 percent of data for training and the last 10 percent of data for testing.
Thus, a model is tested five times on 10 percent of the data, covering the last 30 percent of data over all tests.

\paragraph{Inner Loop: Rolling Origin With Updating But Without Refitting}
The inner loop is repeated for each train-test fold. First, the model is fitted over the train set. The test set evaluation is identical to a rolling origin forecast evaluation with a non-constant holdout size (\citep{adambacktest}) with updating but without refitting of the model. 

Hereby, the forecast origin rolls forward by 1 step at a time, without re-training the model with the new data. This repeats through a single test fold shown in Figure \ref{fig:backtest} until the end of the test fold is reached.
Auto-regressive models relying on lagged observations as inputs are updated with the new data point to produce the next forecast. 
If a model produces multiple forecasts per forecast origin, the average accuracy of all forecast steps is recorded.
For models with unlimited forecast steps which do not depend on being updated with new observations, we forecast over the entire test set at once.

The single-forecast version of this procedure is also known as time series cross-validation in \cite{hyndman2014forecasting}, which we consider unfortunate naming, as it is not to be confused with the outer loop (Expanding Origin Backtest) which is more closely related to $k$-fold cross-validation.

\subsubsection{Metrics}
We calculate metrics on each $i$-step-ahead prediction versus corresponding actual values, and then average this error over all steps.
For evaluation we use two metrics: Mean Absolute Scaled Error (MASE) and  Root Mean Squared Scaled Error (RMSSE), as they are defined in~\cite{mase}. MSSE is given as the RMSE of the evalutated method divided by RMSE of the Na\"ive prediction. This is the squared analogue of mean absolute scaled error (MASE). 
Hereby, Na\"ive refers to predicting the next observation to be identical to the previous value observed.

MASE and RMSSE metrics allow to evaluate the performance of a method compared to Na\"ive forecasting directly, as a value smaller one signifies an improvement over Na\"ive. Further, as the metrics are scaled quantities, different models can be compared across different datasets.

\begin{equation}
    MASE = \frac{\frac{1}{J}\sum_{j=T+1}^{T+J}|y_i - \hat{y}_i|}{\frac{1}{T-1}\sum_{i=2}^{T}|y_i - y_{i-1}|} 
\label{eqn:mase}
\end{equation}

\begin{equation}
    RMSSE = \frac{\sqrt{\frac{1}{J}\sum_{j=T+1}^{T+J}(y_i - \hat{y}_i)^2}}{\sqrt{\frac{1}{T-1}\sum_{i=2}^{T}(y_i - y_{i-1})^2}} 
\label{eqn:msse}
\end{equation}

\begin{align*} 
\textnormal{where} & \\
& T = \textnormal{size of train data} \\
& J = \textnormal{size of test data} 
\end{align*}

\subsection{Computational Resources}
Unless otherwise mentioned, all models were trained and evaluated on a regular laptop with an 8-core CPU (2020 Macbook Pro with M1 chip).

\section{Results}

\subsection{Demonstration of Interpretable Component Decomposition}
The results are displayed separately for experiments S-TS, S-EF, S-TSEF, and S-mTSEF without lagged components and for experiments S-AL, S-TSAL, and S-TSEFAL including lagged components. In the former, we expect both models to perform identically, and in the latter, we expect NeuralProphet to perform better.

\begin{figure}[htbp]
\centering
\begin{subfigure}{.5\textwidth}
  \centering
  \includegraphics[height=0.7\linewidth]{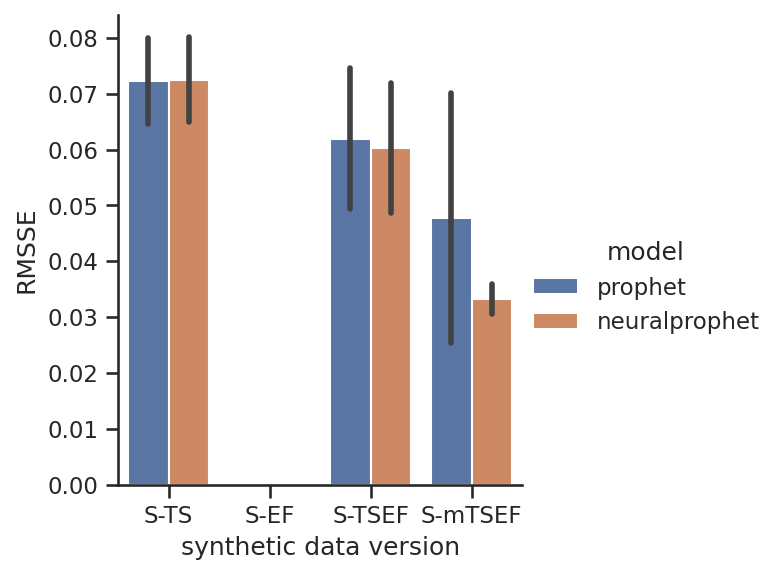} 
  \caption{Experiments without lagged components.}
  \label{fig:no_A_RMSSE}
\end{subfigure}%
\begin{subfigure}{.5\textwidth}
  \centering
  \includegraphics[height=0.7\linewidth]{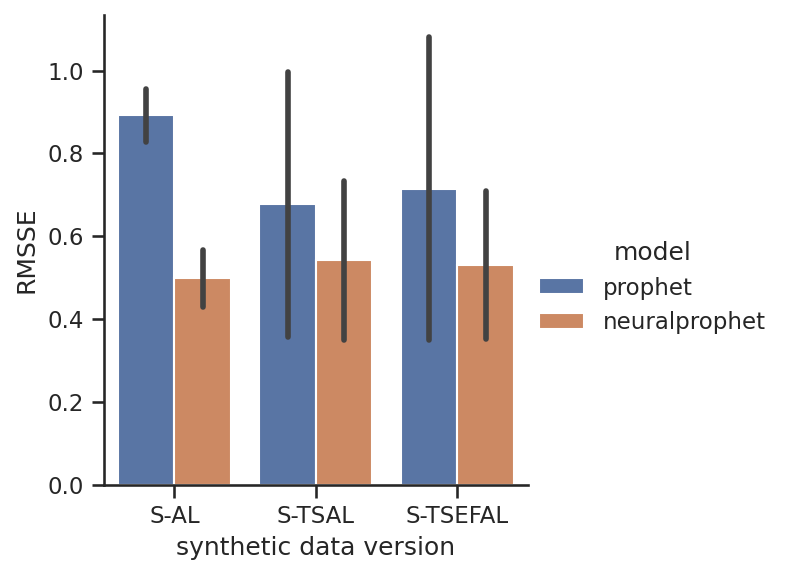}
  \caption{Experiments including lagged components.}
  \label{fig:w_A_RMSSE}
\end{subfigure}
\caption{RMSSE error of the model fit on the synthetic time-series $y$ itself. The plotted bar displays mean value and the line displays standard deviation over 5 runs. }
\label{fig:synth-RMSSE}
\end{figure}

On data without lagged components, NeuralProphet performs identical or slightly better than Prophet, as can be seen in figure~\ref{fig:synth-RMSSE} and figure~\ref{fig:synth-RMSE-noA-comp}. The overall goodness of fit on the 'y' component (the sum of the components) is comparable or slightly better. The biggest difference is seen on \textit{S-mTSEF} with multiplicative components, where NeuralProphet has a significantly lower error and standard deviation of error. 

On experiments involving lagged components, NeuralProphet performs significantly better than Prophet, as seen in figure~\ref{fig:synth-RMSSE} and figure~\ref{fig:synth-RMSE-noA-comp}. Most of the difference comes from a better modelling of lagged regressors.

The fit on the series itself is not as relevant as the component-wise decomposition accuracy, as the overall goodness of fit does not protect from overfitting. The component-wise accuracy, is an adequate measure, as a model that can capture the true underlying dynamics will generalize well.

\subsubsection{Component-Wise Decomposition Accuracy}

\begin{figure}[htbp]
\centering
\begin{subfigure}{.5\textwidth}
    \centering
    \includegraphics[height=0.55\textwidth]{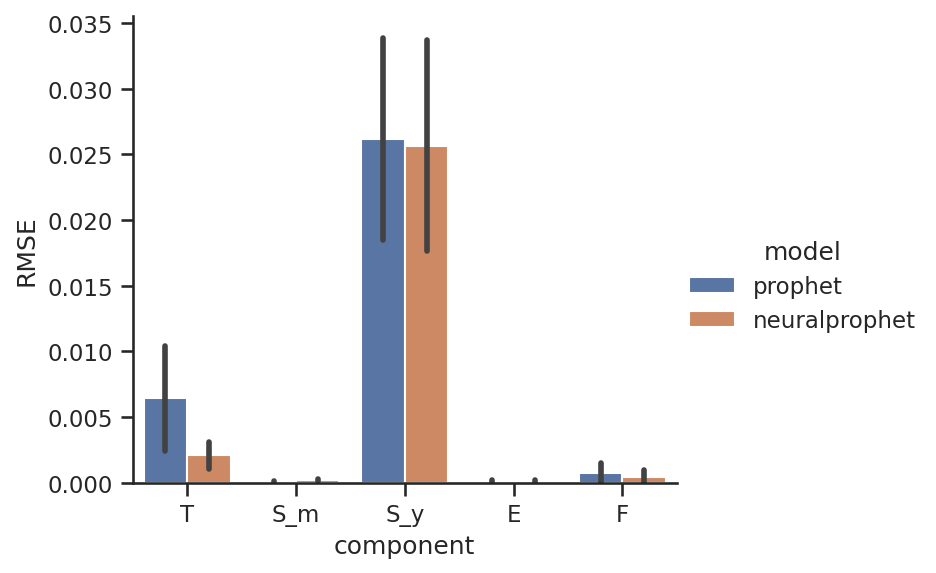}
    \caption{Experiments without lagged components.}
    \label{fig:synth-RMSE-noA-comp}
\end{subfigure}%
\begin{subfigure}{.5\textwidth}
    \centering
    \includegraphics[height=0.55\textwidth]{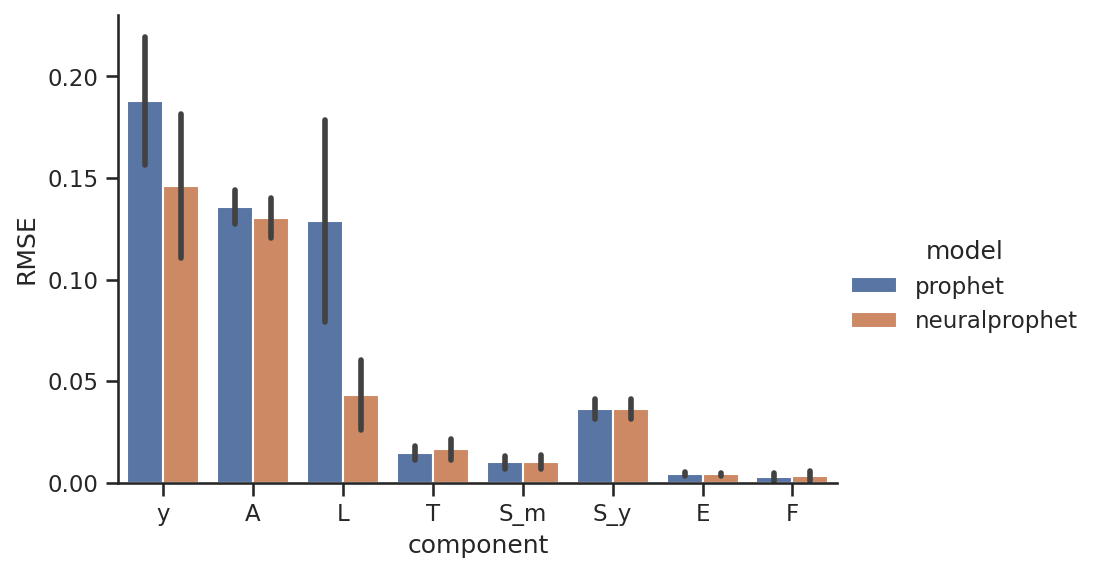}
    \caption{Experiments including lagged components.$^{\star}$}
    \label{fig:synth-RMSE-wA-comp}
\end{subfigure}
    \caption{Component-wise RMSE of predicted forecast component to underlying true generated component value. The plotted bar displays mean value and the line displays standard deviation over all synthetic datasets and over 5 runs. $^{\star}$Note: Prophet does not support Auto-Regression and Lagged Regressor components. Prophet's approximation of said components is implicitly assumed to be zero. Thus, the shown error for Prophet is equivalent to the component's standard deviation.}
    \label{fig:synth-RMSE-comp}
\end{figure}

\paragraph{Without Lagged Components} When compared on the component-wise accuracy NeuralProphet performs identical or better than Prophet, as can be seen in figure~\ref{fig:synth-RMSE-noA-comp}. Both models fit the monthly seasonality, events and future regressors near perfectly. Both models mostly struggle to find a good fit for the yearly seasonality. This is likely explained by the limited number of full annual periods in the data to fit the model on.  A significant difference is observed in the goodness of fit for the trend component, where Neuralprophet exhibits a fraction of the error of Prophet. 

\paragraph{Including Lagged Components} When modelling lagged components, NeuralProphet offers a significant improvement decomposing lagged regressors. However, the AR component is not significantly more accurate than Prophet's zero-prediction. This is due to the fact, the NeuralProphet does not decompose the inputs to the AR module, but rather uses the raw time series values, while in this experiment, we evaluate the performance based on an independently added AR process component. On real applications, such a distinction largely does not matter.

\begin{figure}[htbp]
\centering
\begin{subfigure}[b]{1\textwidth}
    \includegraphics[width=1\linewidth]{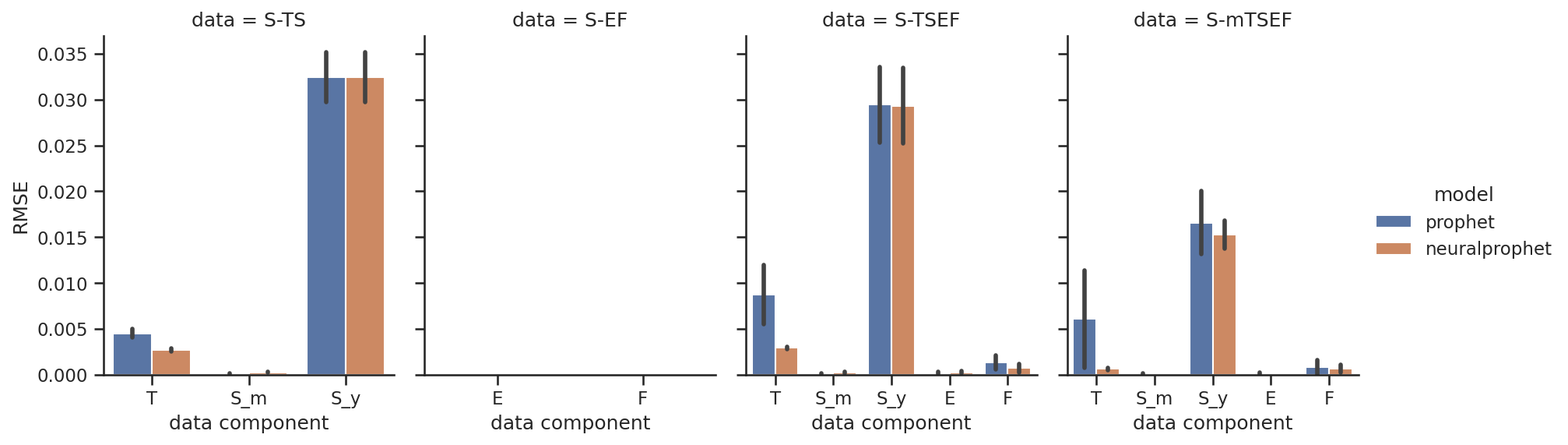}
    \caption{Experiments without lagged components.}
    \label{fig:synth-RMSE-noA}
\end{subfigure}
\begin{subfigure}[b]{1\textwidth}
    \includegraphics[width=1\linewidth]{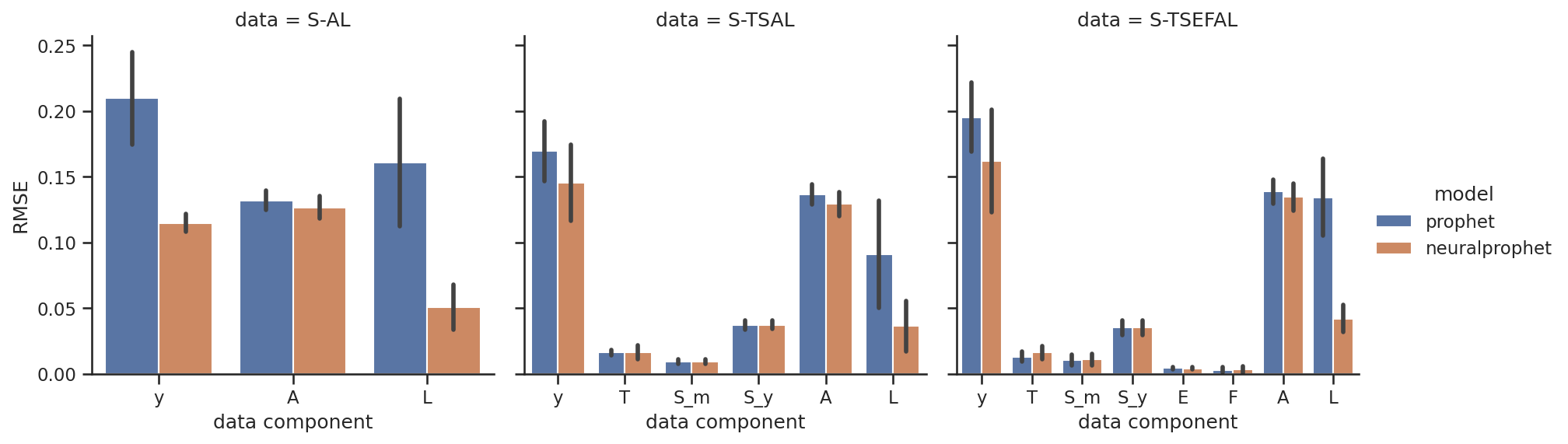}
    \caption{Experiments including lagged components.$^{\star}$}
    \label{fig:synth-RMSE-wA}
\end{subfigure}
    \caption{Here we display the decomposition performance for each experiment and component. Metric is given as RMSE of predicted forecast component to underlying true generated component value. The plotted bar displays mean value and the line displays standard deviation over 5 runs. $^{\star}$Note: Prophet does not support Auto-Regression and Lagged Regressor components. Prophet's approximation of said components is implicitly assumed to be zero. Thus, the shown error for Prophet is equivalent to the component's standard deviation.}
    \label{fig:synth-RMSE}
\end{figure}

Zooming in on the performance for individual experimental setups in Fig.~\ref{fig:synth-RMSE} (Appendix B), we find no notable new observations differing from what we already discussed regarding figure~\ref{fig:synth-RMSSE} and figure~\ref{fig:synth-RMSE-noA-comp}.

\subsubsection{Computational Time}

\begin{figure}[htbp]
    \centering
    \includegraphics[height=0.35\textwidth]{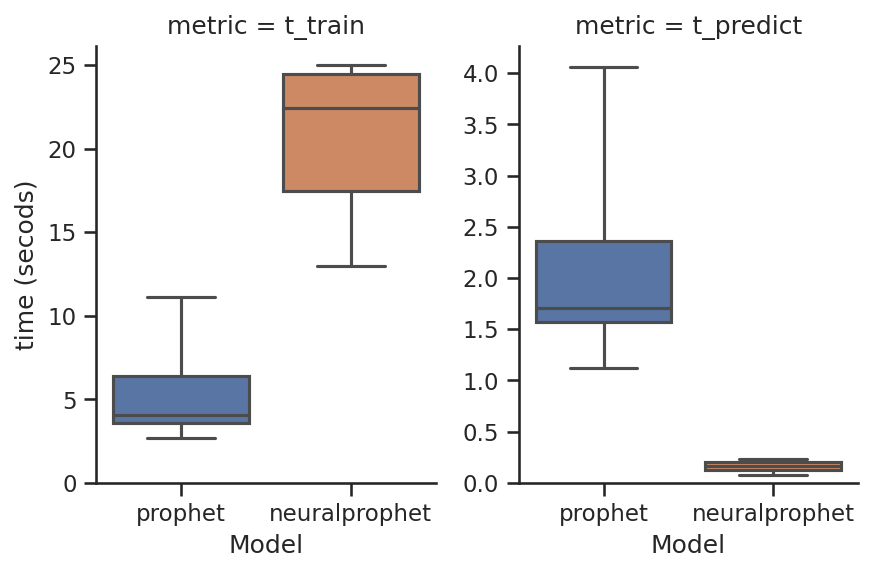} 
    \caption{Training and prediction times across over all folds for experiments without lagged components.}
    \label{fig:synth-time}
\end{figure}

Table~\ref{tab:synth-time} shows that training for NeuralProphet is on average 4.0 times slower compared to Prophet. However, both exhibit a large standard deviation, with Prophet having a long tail upwards and NeuralProphet downwards. 
Prediction time exhibits the inverse relationship, with NeuralProphet computing 13.5 times faster compared to Prophet. NeuralProphet has a tight spread of the prediction times with no upward outliers, while Prophet has a large standard deviation with a long tail upwards, as seen in figure~\ref{fig:synth-time}.

\begin{table}[htbp]
    \centering
\begin{tabular}{lrr}
\toprule
            & Training Time & Prediction Time \\
Model           & (seconds) & (seconds) \\
\midrule
NeuralProphet    &  $20.50~(\pm 4.70)$ & $\textbf{0.16}~(\pm 0.05)$ \\
Prophet           &  $\textbf{5.07}~(\pm 2.30)$ &  $2.16~(\pm 1.08)$\\
\bottomrule
\end{tabular}
    \caption{Computational time in seconds to train a model and time predict over entire dataset. Showing mean and standard deviation over all folds for experiments without lagged components.}
    \label{tab:synth-time}
\end{table}

These observations mean that NeuralProphet may take some extra resources to be fit, but will need significantly less resources in deployment. Further, as NeuralProphet is faster in computing predictions by a magnitude, it makes it possible to deploy it in time-sensitive applications where the next prediction needs to be reliably computed in a fraction of a second.

Thanks to being fully implemented in PyTorch, the NeuralProphet model may be parallelized in the future, allowing it to be deployed on GPUs. We would expect this to close the gap in training time, and the gap in prediction time to widened.

\subsection{Main Benchmark Results}
The overall results are displayed in table~\ref{tab:results_main}. We would like to highlight three observations: First, Prophet and NeuralProphet perform near identical in their default mode. However, both perform significantly worse than Naive one step ahead. Second, NeuralProphet forecasts improve substantially when configuring any amount of lags, consistently performing better than Naive forecasting one step ahead. More lags generally lead to a better performance. Third, when forecasting multiple steps ahead, even when we increase the horizon to 60 steps, NeuralProphet with any number of lags still performs better than Prophet or NeuralProphet in default configuration.

We also observe that the specific number of lags chosen has a small impact on the forecast accuracy, suggesting that the model is not sensitive to optimal parameter choices.

\begin{table}[htbp]
    \centering
\begin{tabular}{lcccc}
\toprule
& \multicolumn{4}{c}{MASE for different forecast horizons} \\
Model &                 1 step &                 3 steps &                 15 steps &                 60 steps \\
\midrule
Prophet$^{\star 1}$         &      \multicolumn{4}{c}{\textbf{8.54} $(\pm 2.17)$}    \\
NeuralProphet$^{\star 1}$    &      \multicolumn{4}{c}{\textbf{8.49} $(\pm 2.03)$}    \\
\midrule
NeuralProphet (1 lags)            &  0.83 $(\pm 0.09)$ &                N/A &                N/A &                N/A \\
NeuralProphet (3 lags)            &  0.72 $(\pm 0.07)$ &                N/A &                N/A &                N/A \\
NeuralProphet (12 lags)           &  \textbf{0.69} $(\pm 0.07)$ &  1.16 $(\pm 0.16)$ &                N/A &                N/A \\
NeuralProphet (30 lags)           &  \textbf{0.62} $(\pm 0.07)$ &  \textbf{0.99} $(\pm 0.12)$ &  \textbf{2.07} $(\pm 0.30)$ &                N/A \\
NeuralProphet (120 lags)$^{\star 2}$        &  \textbf{0.62} $(\pm 0.09)$ &  \textbf{0.94} $(\pm 0.12)$ &  \textbf{1.97} $(\pm 0.29)$ &  \textbf{3.77} $(\pm 0.81)$ \\
\bottomrule
\end{tabular}
    \caption{Forecast error (MASE) for each forecast horizon across all datasets. The standard deviation is shown in paranthesis. The corresponding RMSSE values are given in Table~\ref{tab:rmsse} (Appendix C). Note$^{\star 1}$: The models without lags can produce forecasts for an arbitrary number of forecasts.  Note$^{\star 2}$: Models with 120 lags were not evaluated over monthly datasets due to their short length.}
    \label{tab:results_main}
\end{table}

All model parameters which were set for any of the models are explicitly displayed in Table~\ref{tab:results_main}: number of forecast steps and number of lags. No hyperparameters were manually set or tuned.

The choice of MASE or RMSSE as the error measure is perfectly adequate for the one step ahead forecasting task, but not for multiple steps, as it only measures the naive model's one step ahead forecast error. When forecasting multiple steps ahead, an MASE value of 1 is no longer equal to a naive models performance, as a naive model would likely have a higher error for multiple steps ahead. Nonetheless, the MASE error remains a useful quantity to compare the forecast performance of different models across different datasets, as it scales the error into a somewhat standardized space.

\subsubsection{Ablation Study: Depth of Neural Network}
We further examine the impact of different Neural Network configuration options on the forecast accuracy for different forecast horizons.
Overall, most configurations perform within range of their linear counterparts. Except for the longest forecast horizon of 60 steps, all NN configurations significantly outperform their linear counterpart. 
The specific choices of NN parameters have no discernible impact on the forecast accuracy. Thus, the model is insensitive to non-optimal NN parameter choices, and for short term forecasts, also to lack of any hidden layers.

\begin{table}[htbp]
\small
\centering
\begin{tabular}{lcccc}
\toprule
& \multicolumn{4}{c}{MASE for different forecast horizons} \\
Model &                 1 step &                 3 steps &                 15 steps &                 60 steps \\
\midrule
(30 lags)           &  0.62 $(\pm 0.07)$ &  0.99 $(\pm 0.12)$ &  2.07 $(\pm 0.30)$ &                N/A \\
(30 lags, 1x32 NN)  &  0.60 $(\pm 0.08)$ &  0.93 $(\pm 0.12)$ &  \textbf{1.88} $(\pm 0.23)$ &                N/A \\
(30 lags, 2x24 NN)  &  \textbf{0.59} $(\pm 0.07)$ &  \textbf{0.91} $(\pm 0.13)$ &  1.89 $(\pm 0.29)$ &                N/A \\
(30 lags, 4x16 NN)  &  0.61 $(\pm 0.10)$ &  0.93 $(\pm 0.12)$ &  1.91 $(\pm 0.26)$ &                N/A \\
\midrule
(120 lags)          &  \textbf{0.62} $(\pm 0.09)$ &  0.94 $(\pm 0.12)$ & 1.97 $(\pm 0.29)$ &  3.77 $(\pm 0.81)$ \\
(120 lags, 1x32 NN) &  0.63 $(\pm 0.11)$ &  0.92 $(\pm 0.13)$ &  \textbf{1.81} $(\pm 0.24)$ &  3.06 $(\pm 0.52)$ \\
(120 lags, 2x24 NN) &  0.69 $(\pm 0.14)$ &  \textbf{0.90} $(\pm 0.12)$ &  \textbf{1.81} $(\pm 0.22)$ &  3.07 $(\pm 0.57)$ \\
(120 lags, 4x16 NN) &  0.66 $(\pm 0.11)$ &  0.96 $(\pm 0.15)$ &  1.86 $(\pm 0.31)$ &  \textbf{3.03} $(\pm 0.47)$ \\
\bottomrule
\end{tabular}
    \caption{NeuralProphet (with NN) forecast error (MASE) for each forecast horizon across all datasets. The corresponding RMSSE values are given in Table~\ref{tab:rmsse-nn} (Appendix C). Monthly datasets were excluded due to their insufficient sample sizes to fit the parameters of a NN. The standard deviation is shown in paranthesis. The model definition shows the number of input lags and the number of layers, including their hidden sizes. e.g. '4x16 NN' signifies that the model was set to use 4 hidden layers of dimension 16. Note: For all models with a NN, the learning rate was manually set, but not tuned, to a guesstimate of 0.001.}
    \label{tab:results_NN}
\end{table}

For this ablation study and any other mentions of a configuration involving a NN, the learning rate was manually set to a value of 0.001 for all model variations and datasets. This value is a guesstimate which was not tuned. It was set to avoid potential sources of variation from the learning rate range test. All other model parameters which were set for any of the models are explicitly displayed in Table~\ref{tab:results_NN}: number of forecast steps, number of lags, number of hidden layers, and and hidden layer dimensions.

\subsubsection{Results Based on Data Subject Type}

\begin{table}[htbp] 
\small
\centering
\begin{tabular}{lcccc}
\toprule
 &              human &           economic &        environment &             energy \\
 \midrule
 &            \multicolumn{4}{c}{$\infty$-step  MASE}        \\
\midrule
Prophet                           &  1.11 $(\pm 0.26)$ &  4.25 $(\pm 3.09)$ &  9.27 $(\pm 3.89)$ &  11.13 $(\pm 1.22)$ \\
NeuralProphet                     &  1.13 $(\pm 0.21)$ &  2.39 $(\pm 0.85)$ &  9.48 $(\pm 3.75)$ &  11.13 $(\pm 1.40)$ \\
\toprule
 &            \multicolumn{4}{c}{1-step  MASE}        \\
\midrule
NeuralProphet (1 lags)            &  0.76 $(\pm 0.08)$ &  1.01 $(\pm 0.21)$ &  0.81 $(\pm 0.10)$ &   0.83 $(\pm 0.05)$ \\
NeuralProphet (3 lags)            &  0.75 $(\pm 0.08)$ &  0.98 $(\pm 0.14)$ &  0.76 $(\pm 0.09)$ &   0.63 $(\pm 0.05)$ \\
NeuralProphet (12 lags)           &  0.66 $(\pm 0.07)$ &  0.97 $(\pm 0.17)$ &  0.75 $(\pm 0.08)$ &   0.60 $(\pm 0.04)$ \\
NeuralProphet (30 lags)           &  \textbf{0.62} $(\pm 0.07)$ &  0.82 $(\pm 0.18)$ &  \textbf{0.74} $(\pm 0.10)$ &   0.48 $(\pm 0.03)$ \\
NeuralProphet (120 lags)$^{\star}$          &  0.67 $(\pm 0.15)$ &  \textbf{0.75} $(\pm 0.12)$ &  0.75 $(\pm 0.11)$ &   \textbf{0.46} $(\pm 0.04)$ \\
NeuralProphet (120 lags, 4x16 NN)$^{\star}$ &  0.75 $(\pm 0.13)$ &  0.77 $(\pm 0.06)$ &  0.83 $(\pm 0.16)$ &   0.48 $(\pm 0.07)$ \\
\toprule
 &            \multicolumn{4}{c}{3-step  MASE}        \\
\midrule
NeuralProphet (12 lags)           &  0.77 $(\pm 0.09)$ &  1.81 $(\pm 0.60)$ &  1.25 $(\pm 0.17)$ &  1.10 $(\pm 0.09)$ \\
NeuralProphet (30 lags)           &  \textbf{0.73} $(\pm 0.09)$ &  1.08 $(\pm 0.20)$ &  1.21 $(\pm 0.18)$ &  0.88 $(\pm 0.06)$ \\
NeuralProphet (120 lags)$^{\star}$          &  \textbf{0.73} $(\pm 0.11)$ &  0.98 $(\pm 0.13)$ &  \textbf{1.19} $(\pm 0.18)$ &  0.79 $(\pm 0.06)$ \\
NeuralProphet (120 lags, 4x16 NN)$^{\star}$ &  0.87 $(\pm 0.13)$ &  \textbf{0.83} $(\pm 0.09)$ &  1.27 $(\pm 0.25)$ &  \textbf{0.76} $(\pm 0.08)$ \\
\toprule
 &            \multicolumn{4}{c}{15-step  MASE}        \\
\midrule
NeuralProphet (30 lags)           &  0.89 $(\pm 0.16)$ &  1.58 $(\pm 0.33)$ &  2.52 $(\pm 0.48)$ &  2.17 $(\pm 0.18)$ \\
NeuralProphet (120 lags)$^{\star}$          &  0.93 $(\pm 0.21)$ &  1.30 $(\pm 0.30)$ &  2.45 $(\pm 0.46)$ &  2.04 $(\pm 0.18)$ \\
NeuralProphet (120 lags, 4x16 NN)$^{\star}$ &  \textbf{0.88} $(\pm 0.16)$ &  \textbf{1.18} $(\pm 0.10)$ &  \textbf{2.40} $(\pm 0.57)$ &  \textbf{1.85} $(\pm 0.19)$ \\
\toprule
 &            \multicolumn{4}{c}{60-step  MASE}        \\
\midrule
NeuralProphet (120 lags)$^{\star}$          &  1.38 $(\pm 0.38)$ &  1.70 $(\pm 0.43)$ &  5.14 $(\pm 1.50)$ &  3.78 $(\pm 0.43)$ \\
NeuralProphet (120 lags, 4x16 NN)$^{\star}$ &  \textbf{1.05} $(\pm 0.34)$ &  \textbf{1.52} $(\pm 0.12)$ &  \textbf{3.75} $(\pm 0.81)$ &  \textbf{3.34} $(\pm 0.28)$ \\
\bottomrule
\end{tabular}
\caption{Forecast error (MASE) for each dataset subject type for different forecast horizons. The standard deviation is shown in paranthesis. Note$^{\star}$: Models with 120 lags were not evaluated over monthly datasets due to their short length.}
\label{tab:results_type}
\end{table} 

In general, models with more lags tend to perform better. This is particularly evident in the case of datasets of energy subject type, where the 120-lags NeuralProphet model reduces the forecast error by $96\%$ on 1-step ahead and $66\%$ on 60-step ahead horizons.
Further, we observe that the model configured with a 4 layer, 16 dimension NN is outperformed by linear models on 1 step and partially on 3 step ahead forecasts, while performing best on 15 step and 60 step forecasts for all data types.

Overall, configuring any amount of lags significantly improves the forecasting performance of NeuralProphet for any forecast horizon. Except for 60-step horizons on datasets of human subject type, where only a NN-based model outperforms Prophet and default NeuralProphet.
Measurements of human activity tend to have strong seasonality and have a quickly fading impact of local context. When forecasting into the distant future, statistical patterns tend to dominate human activity data.

\subsection{Task-dependent Strength of Prophet and NeuralProphet}
In Table~\ref{tab:P-NP_comparison} we present our subjective advice on which model to use depending on the task at hand. These are qualitative suggestions based on the empirical evidence we found when evaluating the two models against each other. 

\begin{table}[htbp]
\small
\centering
\begin{tabular}{lcc}
\toprule
Task            & Prophet & NeuralProphet \\
\midrule
Small dataset ($T \ll 100$) & \plusmark & \\
Medium to large dataset ($T \gg 100$) & & \plusmark\\
Long range forecast ($h \gg 100$) & \plusmark & \plusmark\\
Short to medium range forecast ($1 \le h \ll 1000$) & & \plusmark\\
Specific forecast horizon (e.g. $h =24$) & & \plusmark\\
Auto-correlation (dependence on previous observations)  &  & \plusmark \\
Lagged regressors (dependence on observed covariates) & & \plusmark\\
Non-linear dynamics & & \plusmark\\
Global modelling of panel dataset & & \plusmark\\
Frequent retraining on small datasets with   & \multirow{2}{*}{\plusmark} & \\
~~constrained computational resources  & & \\
Fast prediction (computational inference time) & & \plusmark\\
\bottomrule
\end{tabular}
    \caption{Comparison of Prophet's and NeuralProphet's task-dependent strength. \plusmark~marks which of the two models we suggest to use for the given task.}
    \label{tab:P-NP_comparison}
\end{table}

\section{Conclusion}
NeuralProphet is the first hybrid forecasting framework that meets the industry standards for explainability and simplicity-of-use set by Facebook Prophet. As NeuralProphet is based on PyTorch and trained with standard deep learning methods, any algorithm trainable by mini-batch Stochastic Gradient Descent can be included as a module, which makes it easy for developers to extend the framework with new features, and to adopt new research.

NeuralProphet can model local context with auto-regression and covariate regression, making it a suitable tool for time-series where the near-term future depends on the current state of the system. 
This is evidenced by our empirical results where NeuralProphet with auto-regression enabled significantly outperforms Prophet on all forecasting horizons.
Both models perform comparably when configured identically. 
On short to medium-term forecast horizons, most configurations of NeuralProphet reduce the forecast error by $50\%$ to $90\%$. 
Fusing the model with a Neural Network further improves forecast accuracy for medium to long range horizons. 

In our ablation studies, we observe that NeuralProphet's performance is not sensitive to the specific number of lags or the Neural Network configuration. 

NeuralProphet is accessible to forecasting beginners thanks to solid default and automatic hyperparameters, while advanced users can incorporate domain knowledge with optional model customization.
Upgrading Prophet with NeuralProphet empowers forecasting practitioners with an explainable model in a convenient and scalable framework, covering a wide range of forecasting applications.

\section*{Acknowledgements}
We are thankful for the invaluable advice by our academic and industrial advisers at Meta, Facebook AI, Netflix, Uber, Monash and Skoltech; in particular, Italo Lima, Caner Komurlu, Alessandro Panella, Evgeny Burnaev, Sean Taylor, and Benjamin Letham.
We gratefully thank Total Energies for their continued support; in particular, Lluvia Ochoa, Gonzague Henri, and  Michel Lutz.
We would also like to acknowledge the hard work of our lovely open-source contributors; in particular Mateus De Castro Ribeiro, Riley De Haan, and Bernhard Hausleitner.

The work presented herein was funded in part by a research agreement between Stanford University and Total Energies. The views and opinions of authors expressed herein do not necessarily state or reflect those of the funding
source.

\bibliography{refs}

\newpage
\section{Appendices}

\subsection{Appendix A: Details on datasets}
\begin{table}[htbp]
\centering
\begin{tabular}{lllrrr}
\toprule
          Name &        Type &      Subtype &  Freq &      T &  $log_{10}(T)$ \\
\midrule
air\_passengers &       human &      tourism &     M &    144 &            2.2 \\
  retail\_sales &    economic &        sales &     M &    293 &            2.5 \\
peyton\_manning &       human &       clicks &     D &   2905 &            3.5 \\
         birth &       human &   population &     D &   7305 &            3.9 \\
 load\_hospital &      energy &         load &     H &   8760 &            3.9 \\
      solar\_sf & environment &          sun &     H &   8,760 &       3.9 \\
      load\_rte &      energy &         load &     H &  17518 &            4.2 \\
 load\_victoria &      energy &         load & 30min &  17520 &            4.2 \\
yosemite\_temps & environment &      weather &  5min &  18721 &            4.3 \\
         river & environment &        water &     D &  23741 &            4.4 \\
   price\_ercot &    economic & energy price &     H &  25537 &            4.4 \\
   air\_beijing & environment &  air quality &     H &  43800 &            4.6 \\
       sunspot & environment &          sun &     D &  73924 &            4.9 \\
    load\_ercot &      energy &         load &     H & 154872 &            5.2 \\
    power\_wind &      energy &         wind &  1min & 493144 &            5.7 \\
   power\_solar &      energy &        solar &  1min & 493149 &            5.7 \\
\bottomrule
\end{tabular}
\caption{Summary of dataset characteristics. $T$ refers to the dataset length.}
\label{tab:datasets}
\end{table}

\newpage
\subsection{Appendix B: Decomposition RMSE per component per experiment}

\newpage
\subsection{Appendix C: RMSSE per forecast horizon}

\begin{table}[h!] 
\small
\centering
\begin{tabular}{lcccc}
\toprule
& \multicolumn{4}{c}{RMSSE for different forecast horizons} \\
Model &                 1 step &                 3 steps &                 15 steps &                 60 steps \\
\midrule
Prophet$^{\star 1}$            &      \multicolumn{4}{c}{4.37 $(\pm 0.99)$}    \\
NeuralProphet$^{\star 1}$      &      \multicolumn{4}{c}{4.41 $(\pm 0.97)$}    \\
\midrule
NeuralProphet (1 lags)   &  0.71 $(\pm 0.09)$ &                N/A &                N/A &                N/A \\
NeuralProphet (3 lags)   &  0.61 $(\pm 0.08)$ &                N/A &                N/A &                N/A \\
NeuralProphet (12 lags)  &  \textbf{0.59} $(\pm 0.08)$ &  0.92 $(\pm 0.17)$ &                N/A &                N/A \\
NeuralProphet (30 lags)  &  \textbf{0.57} $(\pm 0.11)$ &  \textbf{0.82} $(\pm 0.16)$ &  \textbf{1.51} $(\pm 0.21)$ &                N/A \\
NeuralProphet (120 lags)$^{\star 2}$ &  \textbf{0.51} $(\pm 0.08)$ &  \textbf{0.76} $(\pm 0.11)$ &  \textbf{1.45} $(\pm 0.22)$ &  \textbf{2.46} $(\pm 0.49)$ \\
\bottomrule
\end{tabular}
\caption{Forecast error (RMSSE) for each forecast horizon across all datasets. The standard deviation is shown in paranthesis. Note$^{\star 1}$: The models without lags can produce forecasts for an arbitrary number of forecasts.  Note$^{\star 2}$: The model with 120 lags was not evaluated over monthly datasets due to their short length.}
\label{tab:rmsse}
\end{table}

\begin{table}[h!] 
\small
\centering
\begin{tabular}{lcccc}
\toprule
& \multicolumn{4}{c}{RMSSE for different forecast horizons} \\
Model &                 1 step &                 3 steps &                 15 steps &                 60 steps \\
\midrule
(30 lags, 1x32 NN)  &  0.51 $(\pm 0.07)$ &  0.76 $(\pm 0.11)$ &  1.42 $(\pm 0.20)$ &                N/A \\
(30 lags, 2x24 NN)  &  \textbf{0.50} $(\pm 0.07)$ &  0.75 $(\pm 0.11)$ &  1.43 $(\pm 0.22)$ &                N/A \\
(30 lags, 4x16 NN)  &  0.51 $(\pm 0.08)$ &  0.76 $(\pm 0.11)$ &  1.44 $(\pm 0.21)$ &                N/A \\
(120 lags, 1x32 NN) &  0.52 $(\pm 0.09)$ &  0.75 $(\pm 0.11)$ &  \textbf{1.36} $(\pm 0.20)$ &  \textbf{2.07} $(\pm 0.35)$ \\
(120 lags, 2x24 NN) &  0.55 $(\pm 0.11)$ &  \textbf{0.74} $(\pm 0.11)$ &  1.37 $(\pm 0.19)$ &  2.09 $(\pm 0.39)$ \\
(120 lags, 4x16 NN) &  0.55 $(\pm 0.10)$ &  0.78 $(\pm 0.12)$ &  1.41 $(\pm 0.25)$ &  2.09 $(\pm 0.35)$ \\
\bottomrule
\end{tabular}
\caption{NeuralProphet (with NN) forecast error (RMSSE) for each forecast horizon across all datasets, except monthly, due to their insufficient length to fit a NN. The standard deviation is shown in paranthesis. The model definition shows the number of input lags and the number of layers an their hidden sizes. e.g. 4x16 signifies 4 hidden layers of dimension 16.}
\label{tab:rmsse-nn}
\end{table}

\end{document}